\documentclass{article} 
\usepackage{iclr2026_conference,times}
\iclrfinalcopy

\usepackage{amsmath,amsfonts,bm}

\def\eqref#1{equation~\ref{#1}}

\def\1{\bm{1}}

\DeclareMathAlphabet{\mathsfit}{\encodingdefault}{\sfdefault}{m}{sl}
\SetMathAlphabet{\mathsfit}{bold}{\encodingdefault}{\sfdefault}{bx}{n}

\usepackage[most]{tcolorbox}
\usepackage{xspace}

\usepackage{hyperref}
\usepackage{url}
\usepackage{algorithm}      
\usepackage{algpseudocode} 
\usepackage{todonotes}
\usepackage{amsfonts}  
\usepackage{amssymb}
\usepackage{pifont}
\usepackage{makecell}
\usepackage{ragged2e} 

\usepackage[utf8]{inputenc} 
\usepackage[T1]{fontenc}    
\usepackage{hyperref}       
\usepackage{url}            
\usepackage{booktabs}       
\usepackage{amsfonts}       
\usepackage{nicefrac}       
\usepackage{microtype}      
\usepackage{xcolor}         
\usepackage{xspace}
\usepackage{graphicx}
\usepackage{multicol}
\usepackage{multirow}
\usepackage{colortbl}

\usepackage{colortbl}
\usepackage{booktabs}
\usepackage{setspace}
\usepackage{tcolorbox}

\newtcolorbox{dialogbox}{
  colback=gray!10,  
  boxrule=0pt,      
  arc=0pt,          
  left=10pt,
  right=10pt,
  top=10pt,
  bottom=10pt,
  width=\textwidth,
  boxsep=0pt
}

\usepackage{wrapfig}

\newcommand{\sys}{\mbox{\textsc{MCP-Radar}}\xspace}

\title{\sys: A Multi-Dimensional Benchmark for Evaluating Tool Use Capabilities in Large Language Models}

\author{Xuanqi Gao\thanks{These authors contributed equally to this work.}\\
Xi'an Jiaotong University\\
Xi'an, China\\
\texttt{gxq2000@stu.xjtu.edu.cn} \\
\And
Siyi Xie\(^\ast\) \\
Xi'an Jiaotong University\\
Xi'an, China\\
\texttt{xsy2225025620@stu.xjtu.edu.cn} \\
\AND
Juan Zhai\\
University of Massachusetts at Amherst\\
Amherst, USA\\
\texttt{juanzhai@umass.edu} \\
\And
Shiqing Ma\\
University of Massachusetts at Amherst\\
Amherst, USA\\
\texttt{shiqingma@umass.edu} \\
\And
Chao Shen\\
Xi'an Jiaotong University\\
Xi'an, China\\
\texttt{chaoshen@mail.xjtu.edu.cn}
}

\begin{document}

\maketitle

\begin{abstract}
As Large Language Models (LLMs) evolve from passive text generators to active reasoning agents capable of interacting with external tools, the Model Context Protocol (MCP) has emerged as a key standardized framework for dynamic tool discovery and orchestration. Despite its widespread industry adoption, existing evaluation methods do not adequately assess tool utilization capabilities under this new paradigm.
To address this gap, this paper introduces \sys, the first comprehensive benchmark specifically designed to evaluate LLM performance within the MCP framework. \sys features a challenging dataset of 507 tasks spanning six domains: mathematical reasoning, web search, email, calendar, file management, and terminal operations. It quantifies performance based on two primary criteria: answer correctness and operational accuracy. To closely emulate real-world usage, our evaluation employs both authentic MCP tools and high-fidelity simulations of official tools.
Unlike traditional benchmarks that rely on subjective human evaluation or binary success metrics, \sys adopts objective, quantifiable measurements across multiple task domains, including computational resource efficiency and the number of successful tool-invocation rounds. Our evaluation of leading closed-source and open-source LLMs reveals distinct capability profiles and highlights a significant trade-off between accuracy and efficiency.
Our findings provide actionable insights for both LLM developers and tool creators, establishing a standardized methodology applicable to the broader LLM agent ecosystem. All implementations, configurations, and datasets are publicly available at \url{https://anonymous.4open.science/r/MCPRadar-B143}.
\end{abstract}

\section{Introduction}\label{sec:intro}

The paradigm of Large Language Models (LLMs) is undergoing a fundamental shift, evolving from passive text generators into proactive reasoning agents capable of interacting with external tools and APIs~\citep{chowdhery2022palmscalinglanguagemodeling,brown2020languagemodelsfewshotlearners}. This evolution has been significantly accelerated by the advent of the Model Context Protocol (MCP), which provides a standardized framework for dynamic tool discovery and orchestration~\citep{IntroducingModelContext,ModelContextProtocol,QwenLMQwenAgent2025}. As MCP adoption becomes widespread, the development of rigorous, standardized benchmarks to evaluate model performance within this new paradigm is critically important.

However, existing evaluation methodologies are insufficient. While traditional benchmarks excel at assessing knowledge-based reasoning~\citep{hendrycksMeasuringMassiveMultitask2021}~\citep{zhongAGIEvalHumanCentricBenchmark2023} or instruction following~\citep{wangMINTEvaluatingLLMs2024}~\citep{shridharALFWorldAligningText2021}, they offer limited insight into tool-use capabilities. Current tool-centric evaluations suffer from two primary limitations: 1) they struggle to differentiate between a model's genuine problem-solving via tools and mere recitation of pre-trained knowledge, 2) their reliance on simulated environments often fails to capture the complexities of real-world tool interactions.

\begin{figure}[h]
    \centering
    \includegraphics[trim={2.1cm 16cm 1.7cm 6.15cm},clip,width=\linewidth]{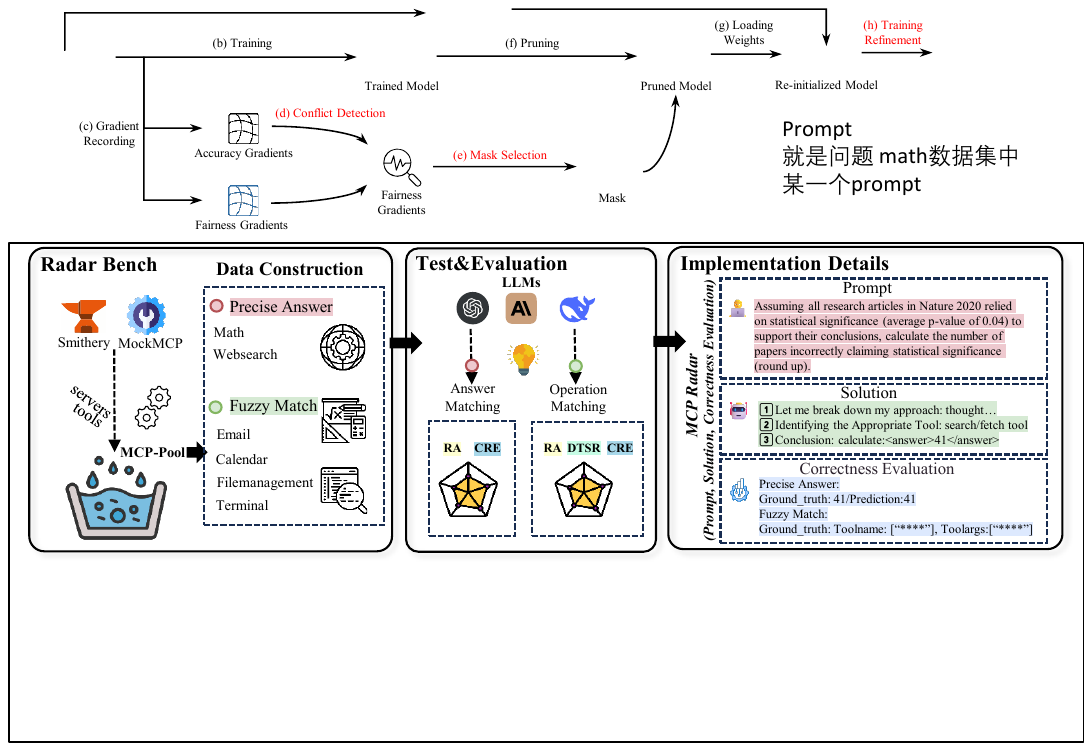}
    
    \caption{Overview of {\sys}.}
    \label{fig:overview}
\end{figure}

To address these gaps, we introduce \sys, the first comprehensive benchmark designed specifically to evaluate LLM performance in the MCP paradigm. As illustrated in \autoref{fig:overview}, our methodology is structured around three core stages. First, in the Data Construction phase, we curate a diverse MCP Pool using both real-world tools from platforms like Smithery and high-fidelity mock MCPs for common applications such as email and calendar management. This pool supports two distinct task categories: Precise Answer tasks (e.g., Math, Websearch), which have a single correct ground-truth value, and Fuzzy Match tasks (e.g., Filemanagement, Terminal), which require a correct sequence of operations.

Next, in the Test \& Evaluation stage, we evaluate ten leading open- \& closed-source LLMs. Our novel framework moves beyond simple binary success metrics by assessing accuracy through two core methods: Answer Matching for Precise Answer tasks and Operation Matching for Fuzzy Match tasks. Based on this, we quantify performance across multiple dimensions: Answer Accuracy (\textbf{RA}), Tool Selection Efficiency (\textbf{DTSR}), and Computational Resource Efficiency (\textbf{CRE}).

Our evaluation using \sys reveals critical insights. For instance, while closed-source models significantly outperform open-source counterparts in mathematical reasoning, this gap narrows to less than 10\% in web search tasks. More importantly, we identify a recurring failure pattern: models frequently select a semantically plausible but functionally incorrect tool, indicating a superficial understanding of the task requirements. Based on these findings, we provide actionable recommendations for both LLM development and MCP tool design.

    

Our main contributions are threefold:
\begin{itemize}
\item We introduce \sys, a comprehensive benchmark featuring two fundamental task types (Precise Answer and Fuzzy Match) across six critical domains: Mathematical Reasoning, Web Search, Email, Calendar, File Operations, and Terminal.
\item We establish a high-fidelity evaluation environment by employing a combination of real-world MCP tools and meticulously replicated mock tools based on official specifications.
\item We propose a novel, multi-dimensional evaluation framework for tool-augmented LLMs that utilizes purely objective and quantifiable metrics to assess accuracy, efficiency, and resourcefulness.
\end{itemize}

\section{Related work}\label{sec:rw}

\subsection{The Evolution Towards Standardized Tool Use}

The Model Context Protocol (MCP) for Large Language Models (LLMs) is a unified interaction standard proposed by Anthropic to address systematic challenges in tool invocation. 
Early LLMs such as the GPT family relied only on static training data and were unable to access real-time information or interact with external systems, resulting in limited applications. 
~\citet{wei2022chain} demonstrated the role of structured reasoning in improving the performance of LLMs. 
With the increase of complex scenarios such as multi-round dialog systems, developers try to connect to external via API~\citep{liu2024autofeedback}~\citep{song2023restgpt}~\citep{qin2023toolllm}~\citep{tang2023toolalpaca} to external data sources. 
Use the tool-enhanced LLM~\citep{patil2024gorilla}~\citep{parisi2022talm}~\citep{lu2023chameleon} to try to solve the web browsing ~\citep{schick2020exploiting}~\citep{ spiegelwebmap}~\citep{chowdhury2024ai} or code interpretation ~\citep{zhuang2023toolqa}~\citep{liu2023agentbench} and other aspects of relevance, but Schick et al.~\citep{ schick2023toolformer} points out that this ``peer-to-peer'' integration leads to \(N\times M\) issues, limiting system expansion and increasing maintenance costs. 
While platforms such as Hugging Face promote model sharing, and frameworks such as LangChain attempt to enhance model capabilities through Tool Calling, these solutions do not address the underlying problem. However, these solutions still fail to address the fundamental problem. 
~\citet{hsieh2023tool} noting that these approaches still lack a common context delivery mechanism. 
It is in this context that the Model Context Protocol (MCP) was formally introduced and open sourced.


\subsection{Evaluating Tool and MCP Proficiency}
Evaluating the tool-use capabilities of LLMs has emerged as a critical research direction. While traditional evaluation frameworks focused on language comprehension, the advent of tool-augmented AI \citep{wang2023interactive,schick2023toolformer} has made specialized benchmarks for tool proficiency essential \citep{xu2023tool,liang2024taskmatrix,patil2024gorilla}. The HELM framework, proposed by \citet{liang2022holistic}, sought to establish multi-dimensional evaluation criteria but did not specifically address the efficiency and effectiveness of tool interaction protocols. Existing tool-use benchmarks exhibit several limitations: some struggle to handle complex scenarios such as long-context memory, multi-turn, or multi-tool calls \citep{li2023api,patil2024gorilla,xu2023tool,zhuang2023toolqa,tang2023toolalpaca,qin2023toolllm}, while others rely on single-path, standardized answers that do not align with the diversity of real-world user needs \citep{wang2024mtu}. Furthermore, the dataset in \citet{luo2025mcpuniversebenchmarkinglargelanguage} is limited to operational-matching tasks, and the one in \citet{liu2025mcpevalautomaticmcpbaseddeep} is entirely synthetic, raising doubts about its real-world applicability. Consequently, these systems lack a comprehensive and systematic evaluation of a model's ability to utilize specific protocols like MCP. In this paper, we introduce \sys , a large-scale instruction benchmark, to explore the performance of LLMs in a variety of real-world MCP usage scenarios.




\section{\sys Data Generation}\label{sec:design}
\label{sec:dataset}
The \sys benchmark is comprised of 507 instances meticulously crafted to span six distinct real-world domains. To comprehensively evaluate the diverse capabilities of LLM agents, we structured our dataset around two fundamental task archetypes: Precise Answer and Fuzzy Match. A detailed breakdown of the instance distribution and the specific tools associated with each domain is provided in \autoref{tab:domains}.

\begin{figure}[h]
    \centering
    \scalebox{0.9}{
    \includegraphics[trim={1.63cm 16.5cm 1.76cm 7.5cm},clip,width=\linewidth]{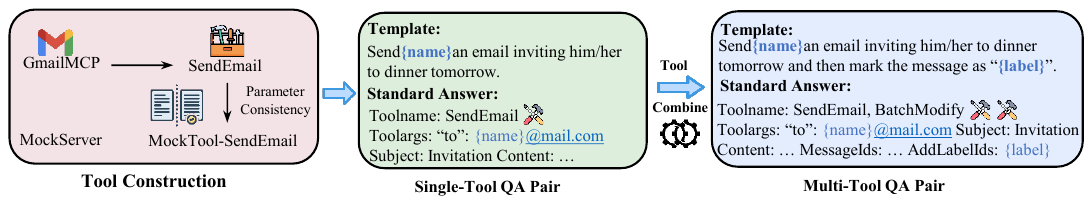}}
    
    \caption{Data Generation.}
    \label{fig:datagene}
\end{figure}

The two task categories are defined as follows:
\textbf{Precise Answer:} This category includes tasks that require the model to return a single, definitive ground-truth value, such as a number, an algebraic expression, or a specific noun. As detailed in \autoref{tab:domains}, this category covers the Math and Websearch domains. To ensure robustness and relevance, the instances for these tasks were curated and adapted from established academic datasets. Each data point consists of a query and its unique, verifiable answer. \textbf{Fuzzy Match:} This category encompasses operational tasks where success is determined not by a simple textual response, but by the correct invocation of an external tool with the appropriate parameters. This is essential for evaluating an agent's ability to act upon instructions in domains like Email, Calendar, Filemanagement, and Terminal. For these tasks, each data point consists of a query paired with the ground-truth tool name and its corresponding arguments. The step-by-step methodology for generating these goal-oriented instances is illustrated in \autoref{fig:datagene}, with concrete examples available in \autoref{app:0}.


\begin{table}[!h]
    \centering
    \footnotesize
    \scalebox{0.9}{
    \begin{tabular}{lcccc}
        \toprule
        \textbf{Task Type} & \textbf{Data-Domain} & \textbf{Quantity} & \textbf{\#Tools} & \textbf{Tools}\\
        \midrule
        \multirow{2}{*}{\centering Precise Answer} & Math & 120 & 4 & Calculate, SolveEquation, Differentiate...\\
        & Websearch & 94 & 2 & Search, FetchContent \\
        \multirow{4}{*}{\centering Fuzzy Match} & Email & 119 & 17 & SendEmail, DraftEmail, ReadEmail... \\
        & Calendar & 28 & 4 & ListCalendars, ListEvents, CreateEvent...\\
        & Filemanagement & 91 & 13 & ReadTextFile, ReadMediaFile, ReadMultipleFiles... \\
        & Terminal & 63 & 9 & GetConfig, SetConfigValue, StartProcess...\\
        \bottomrule
    \end{tabular}
    }
    \caption{Data-Tool Statistics.}
    \label{tab:domains}
\end{table}

\subsection{Precise Answer Data Curation}

Our methodology for the Precise Answer dataset prioritizes answer accuracy, real-world relevance, and challenging queries. To achieve this, we chose to adapt and repurpose existing authoritative datasets for mathematics and web search, rather than generating synthetic data from scratch~\citep{zhuang2023toolqa} or deriving tasks solely from tool definitions~\citep{styles2024workbenchbenchmarkdatasetagents}. This approach grounds our benchmark in previously validated problems.

The curation process involved several key steps:
\begin{enumerate}
    \item Data Sourcing and Filtering: We began by selecting the most challenging queries from high-quality source datasets~\citep{gou2024toratoolintegratedreasoningagent,fan2024hardmathbenchmarkdatasetchallenging,srivastava2023imitationgamequantifyingextrapolating,kazemi2025bigbenchextrahard,mialonGAIABenchmarkGeneral2023}. To ensure our benchmark specifically tests tool-use rather than a model's internal knowledge—a common issue of data contamination—we used a powerful baseline model (Gemini 2.5 Flash~\citep{Gemini25Pro}) as a filter. Queries that the model could solve without external tools were discarded, isolating problems that genuinely necessitate tool invocation.
    \item Ground-Truth Annotation: For the remaining queries, we manually annotated the scope of potentially applicable tools for each problem. This annotation defines a set of valid tools without enforcing a single, rigid solution path. Crucially, for this task category, our evaluation focuses solely on the correctness of the final standard answer; the specific tools used are not assessed, only the result.
    \item Tool Implementation: To execute these tasks, we integrated verified, open-source MCP tools. Specifically, we utilized the calculator-mcp-server for mathematical operations and the duckduckgo-mcp-server for web search functionalities.
\end{enumerate}

\subsection{Fuzzy Match Data Curation}

The generation of the Fuzzy Match dataset involved two core stages: establishing a controlled tool environment and programmatically generating single- and multi-tool interaction scenarios.

\paragraph{Tool Implementation and Environment}
To ensure reproducible and monitorable experiments, we developed a high-fidelity, semi-sandboxed tool environment. For the Filemanagement and Terminal domains, we integrated robust, community-developed open-source MCP tools. For the Email and Calendar domains, we implemented our own mock tools, EmailMCP and CalendarMCP, which meticulously replicate the interfaces and parameter structures of their real-world counterparts (GmailMCP and macOS CalendarMCP, respectively). These mock tools interact with a controlled, local database, pre-populated with 100 email and 50 calendar entries, rather than executing live operations. This setup provides realistic tool interaction schemas while maintaining a controlled evaluation environment. The email data format is detailed in \autoref{app:0}.

\paragraph{Instance Generation Methodology}
Our approach employs a template-based programmatic method, similar to frameworks like Workbench~\citep{styles2024workbenchbenchmarkdatasetagents} and ToolQA~\citep{zhuang2023toolqa}, to generate question-and-action pairs.



\begin{wrapfigure}{R}{0.4\textwidth}
    \centering
    \scalebox{1}{
    \includegraphics[trim={0cm 0cm 0cm 0cm},clip,width=\linewidth]{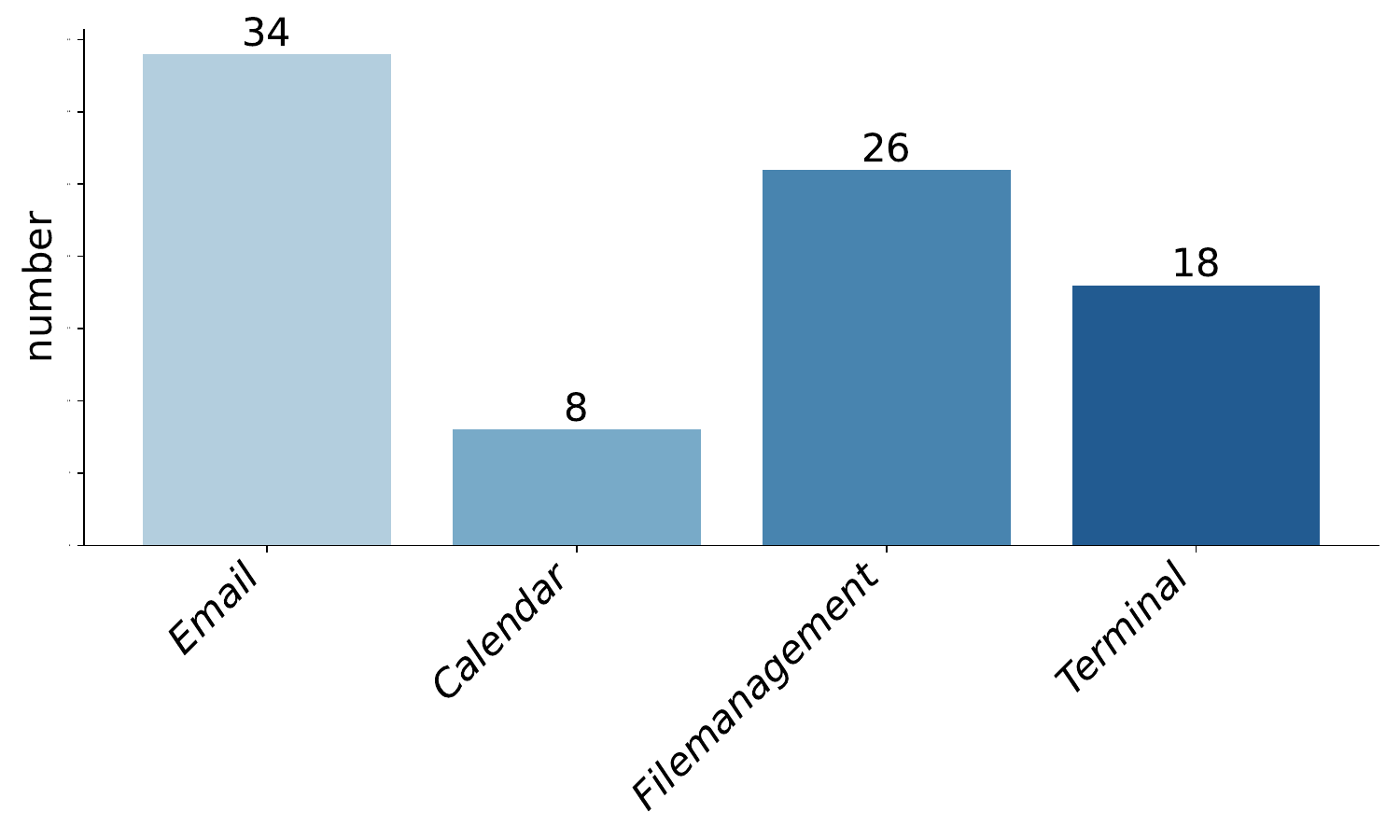}
    }
    \caption{Multi-Tool Data Distribution.}
    \label{fig:digital}
\end{wrapfigure}

For single-tool instances, we designed a unique template for each tool, from which five distinct tasks were generated. This process co-generates both the user query and the corresponding ground-truth tool invocation (toolname and toolargs), which forms the basis for our Fuzzy Match evaluation. Examples of these templates are provided in \autoref{app:0}.

For multi-tool instances, we adopted a more constrained approach to avoid the combinatorial explosion of exhaustive tool pairings. We identified the top three most frequently used tools within each domain and created chained-task templates\autoref{app:0} by combining their respective single-tool templates. To ensure a unique and verifiable solution, the sequence of tool invocations in these multi-tool scenarios is strictly defined in the ground truth. The distribution of multi-tool questions per domain is shown in \autoref{fig:digital}.

\section{Experiment}\label{sec:eval}

\subsection{Experimental Setup}

\paragraph{Models Evaluated}
Our evaluation encompasses a diverse suite of ten leading Large Language Models, accessed via the OpenRouter API for standardized interfacing. The selection includes six state-of-the-art closed-source models: openai/gpt-5~\citep{GPT5}, openai/gpt-4o~\citep{HelloGPT4o}, google/gemini-2.5-flash~\citep{Gemini25Flash}, google/gemini-2.5-pro~\citep{Gemini25Pro}, anthropic/claude-sonnet-4~\citep{ClaudeSonnet4}, and anthropic/claude-3.7-sonnet~\citep{Claude37Sonnet}; and four prominent open-source models: qwen/qwen3-235b-a22b-2507~\citep{Qwen3}, deepseek/deepseek-chat-v3-0324~\citep{yangDeepSeekV3Advanced2024}, deepseek/deepseek-r1-0528~\citep{DeepSeekR1}, and meta-llama/llama-4-maverick~\citep{Llama4Maverick}.

\paragraph{Implementation Details}
Each model was tasked with solving problems using a set of 49 MCP tools distributed across the six domains. For every task, the model was provided with a system\_message, the user question, and a list of available tools. The models were required to complete each task within a maximum of K=10 interaction rounds; exceeding this limit was considered a task failure.

The required output format depended on the task type. For Fuzzy Match tasks, the evaluation focused on the sequence of tool invocations generated by the model. For Precise Answer tasks, models were instructed to enclose their final response within a designated tag: <answer>[YOUR FINAL ANSWER]</answer>, ensuring unambiguous extraction of the answer. To mitigate tool-related hallucinations~\citep{Huang_2025} and improve reliability, the system\_message included detailed, tool-specific instructions. The complete prompt templates, along with a comparative analysis of different prompting strategies (e.g., ReAct vs. concise), are available in \autoref{app:1}.

\subsection{Evaluation Metrics}

Our evaluation framework employs distinct sets of metrics tailored to the unique success criteria of each task category.

\paragraph{Precise Answer Tasks}
For this category, evaluation focuses exclusively on the final outcome, as the tool-use path to a correct answer is often non-unique and may involve self-correction from erroneous steps. Consequently, we do not assess the intermediate tool invocation sequence. Performance is measured along two dimensions:

\begin{itemize}
\item \textbf{Result Accuracy (RA):} A binary metric indicating whether the model's final, extracted answer matches the ground truth exactly.
\item \textbf{Computational Resource Efficiency (CRE):} A measure of the computational cost (e.g., token consumption) incurred to reach the solution.
\end{itemize}

We only require that the tools used by the model are from the valid set provided for the task, but we do not penalize alternative or redundant tool paths as long as the final answer is correct.

\paragraph{Fuzzy Match Tasks}
For tasks where the goal is to perform a correct operation, we evaluate the tool invocation process itself. Performance is assessed across three dimensions:

\begin{itemize}
\item \textbf{Result Accuracy (RA):} A binary metric indicating whether the model's final, decisive tool invocation (both tool name and parameters) exactly matches the ground-truth operation.
\item \textbf{Dialogue Turn Success Rate (DTSR):} Defined as the ratio of successful tool invocations to the total number of interaction turns. A "successful invocation" is one where the model selects an applicable tool and provides correctly formatted parameters, measuring its step-by-step ability to extract information and construct valid calls.
\item \textbf{Computational Resource Efficiency (CRE):} A measure of the computational cost, with values normalized using max-min scaling to allow for cross-model comparison.
\end{itemize}

\section{Results }\label{sec:results}

\subsection{Main Results}

\paragraph{Precise Answer Tasks}

As shown in \autoref{tab:main1}, closed-source models generally exhibit superior performance in this category. The performance gap is most pronounced in mathematical reasoning. The Websearch domain proved to be highly challenging for all models, with success rates universally below 30\%. This difficulty stems from the dual requirement of selecting the correct tool and formulating a precise query to extract the necessary information. Among the models tested, Gemini-2.5-Pro emerged as the top performer with an accuracy of 29.8\%, whereas the open-source Llama-4 recorded the lowest at 0.8\%. While closed-source models maintained an advantage in Websearch, the performance gap narrowed compared to other tasks, with an average success rate of 20.7\% versus 10.8\% for open-source models.

\begin{table}[!]
\caption{Comparison of model performance metrics across two domains of \sys. Gemini-Flash and Gemini-Pro are based on Gemini 2.5 Flash and Gemini 2.5 Pro respectively. Scores highlighted in \colorbox[HTML]{FBDED0}{red} indicate the lowest score, while scores in \colorbox[HTML]{D1E5D9}{green} are the highest.}\label{tab:main1}
\centering
\footnotesize
\scalebox{0.66}{
\begin{tabular}{l|l|rrrrrrrrrr}
\toprule
\multicolumn{2}{c|}{\textbf{Task}} & \multicolumn{1}{c}{\textbf{GPT-5}} & \multicolumn{1}{c}{\textbf{GPT-4o}} & \multicolumn{1}{c}{\textbf{Gemini-Flash}} & \multicolumn{1}{c}{\textbf{Gemini-Pro}} & \multicolumn{1}{c}{\textbf{Claude-3.7}} & \multicolumn{1}{c}{\textbf{Claude-4}} & \multicolumn{1}{c}{\textbf{Qwen3}} & \multicolumn{1}{c}{\textbf{Deepseek-V3}} & \multicolumn{1}{c}{\textbf{Deepseek-R1}} & \multicolumn{1}{c}{\textbf{Llama-4}} \\
\midrule
\multirow{2}{*}{Math} & \textbf{ACC.} & 0.607 & 0.486 & \cellcolor[HTML]{D1E5D9}0.612 & 0.539 & 0.466 & 0.423 & 0.408 & 0.287 & 0.365 & \cellcolor[HTML]{FBDED0}0.128 \\
                      & \textbf{CRE.} & 0.564 & 0.326 & 0.688 & 0.403 & 0.456 & \cellcolor[HTML]{FBDED0}0.000 & 0.785 & 1.000 & 0.965 & 0.644 \\

\midrule
\multirow{2}{*}{Websearch} & \textbf{ACC.} & 0.182 & 0.154 & 0.193 & \cellcolor[HTML]{D1E5D9}0.298 & 0.256 & 0.164 & 0.194 & 0.103 & 0.125 & \cellcolor[HTML]{FBDED0}0.008 \\
                           & \textbf{CRE.} & 0.245 & \cellcolor[HTML]{FBDED0}0.000 & 0.421 & 0.324 & 0.231 & 0.364 & 0.764 & 0.897 & \cellcolor[HTML]{D1E5D9}1.000 & 0.965 \\
\bottomrule
\end{tabular}
}
\end{table}


\paragraph{Fuzzy Match Tasks}
In this category, model performance strongly correlates with task complexity. Models achieved significantly higher accuracy on simpler operational tasks (Email, Calendar) compared to more complex domains requiring precise sequential logic (Filemanagement, Terminal). A critical finding, detailed in \autoref{tab:main2}, is the significant disparity observed between Dialogue Turn Success Rate (DTSR) and final accuracy (ACC) in complex tasks. For instance, in the Filemanagement domain, GPT-4o achieved a high DTSR of 84.5\% but an ACC of only 43\%. This 40.7\% gap highlights a crucial failure mode: models can syntactically execute tool calls correctly but fail to select the semantically appropriate tool to solve the problem. This suggests a superficial understanding of the task's core requirements.

\begin{table}[!]
\caption{Comparison of model performance metrics across Four domains of \sys}\label{tab:main2}
\centering
\footnotesize
\scalebox{0.63}{
\begin{tabular}{l|l|rrrrrrrrrr}
\toprule
\multicolumn{2}{c|}{\textbf{Task}} & \multicolumn{1}{c}{\textbf{GPT-5}} & \multicolumn{1}{c}{\textbf{GPT-4o}} & \multicolumn{1}{c}{\textbf{Gemini-Flash}} & \multicolumn{1}{c}{\textbf{Gemini-Pro}} & \multicolumn{1}{c}{\textbf{Claude-3.7}} & \multicolumn{1}{c}{\textbf{Claude-4}} & \multicolumn{1}{c}{\textbf{Qwen3}} & \multicolumn{1}{c}{\textbf{Deepseek-V3}} & \multicolumn{1}{c}{\textbf{Deepseek-R1}} & \multicolumn{1}{c}{\textbf{Llama-4}} \\
\midrule
\multirow{3}{*}{Email} & \textbf{ACC.} & 0.749 & 0.632 & 0.742 & \cellcolor[HTML]{D1E5D9}0.825 & 0.765 & \cellcolor[HTML]{FBDED0}0.454 & 0.756 & 0.625 & 0.738 & 0.448 \\

                      & \textbf{CRE.} & 0.642 & \cellcolor[HTML]{D1E5D9}1.000 & 0.854 & 0.765 & 0.846 & 0.213 & 0.413 & 0.632 & 0.875 & \cellcolor[HTML]{FBDED0}0.000 \\
                      & \textbf{DTSR.} & 0.806 & 0.765 & \cellcolor[HTML]{D1E5D9}0.936 & 0.855 & 0.784 & 0.645 & 0.802 & 0.743 & 0.932 & \cellcolor[HTML]{FBDED0}0.623 \\
\midrule
\multirow{3}{*}{Calendar} & \textbf{ACC.} & 0.723 & 0.643 & 0.762 & \cellcolor[HTML]{D1E5D9}0.825 & 0.765 & 0.653 & 0.746 & 0.432 & 0.312 & \cellcolor[HTML]{FBDED0}0.286 \\
                           & \textbf{CRE.} & 0.423 & 0.325 & 0.412 & 0.333 & 0.531 & \cellcolor[HTML]{FBDED0}0.000 & 0.352 & \cellcolor[HTML]{D1E5D9}1.000 & 0.862 & 0.742 \\
                           & \textbf{DTSR.} & 0.802 & 0.695 & 0.783 & \cellcolor[HTML]{D1E5D9}0.886 & 0.823 & 0.663 & 0.769 & 0.502 & 0.612 & \cellcolor[HTML]{FBDED0}0.466 \\
\midrule
\multirow{3}{*}{Filemanagement} & \textbf{ACC.} & 0.323 & 0.438 & 0.346 & \cellcolor[HTML]{D1E5D9}0.596 & 0.462 & 0.436 & 0.478 & 0.362 & 0.392 & \cellcolor[HTML]{FBDED0}0.254 \\
                      & \textbf{CRE.} & 0.000 & 0.432 & 0.852 & 0.754 & 0.532 & 0.756 & \cellcolor[HTML]{FBDED0}0.412 & 0.751 & 0.651 & \cellcolor[HTML]{D1E5D9}1.000 \\
                      & \textbf{DTSR.} & 0.522 & \cellcolor[HTML]{D1E5D9}0.845 & 0.543 & 0.623 & 0.588 & 0.623 & 0.563 & \cellcolor[HTML]{FBDED0}0.452 & 0.753 & 0.635 \\
\midrule
\multirow{3}{*}{Terminal} & \textbf{ACC.} & 0.420 & 0.413 & 0.562 &\cellcolor[HTML]{D1E5D9} 0.599 & 0.458 & 0.396 & 0.452 &\cellcolor[HTML]{FBDED0} 0.325 & 0.366 & 0.421 \\
                      & \textbf{CRE.} & 0.233 & 0.153 & 0.425 & \cellcolor[HTML]{D1E5D9}1.000 & 0.624 & 0.356 & 0.451 & 0.222 & 0.346 & \cellcolor[HTML]{FBDED0}0.000 \\
                      & \textbf{DTSR.} & 0.453 & 0.566 & 0.608 & 0.665 & 0.496 & 0.455 & 0.469 & \cellcolor[HTML]{FBDED0}0.365 &\cellcolor[HTML]{D1E5D9} 0.666 & 0.652 \\
\bottomrule
\end{tabular}
}
\end{table}

To further probe the models' planning capabilities, we conducted a targeted multi-tool experiment inspired by~\citet{huang2023metatool}. We tested whether providing a hint about the number of required tools would improve performance. The results in \autoref{tab:toolchoose} show that such prompts had minimal impact, yielding only a 2.5\% to 5\% improvement. This indicates that the primary capability boundary for current models is not determining if a tool is needed, but rather deciding which specific tool to invoke and how to parameterize it correctly.

\begin{table}[!]
\caption{Multi Tool Selection Result. 2/2 means that we suggest call two tools, and then the LLM call the correct two tools. 2/ means that we do not advise about tools, and the LLM called the correct two tools. 1/2 means that on the basis of suggestion, the LLM called two tools, but only one of them was correct. 1/1 means that, based on the suggestion, the LLM only call one tool and it is correct}\label{tab:toolchoose}

\centering
\footnotesize
\scalebox{0.7}{
\begin{tabular}{l|l|rrrrrrrrrr}
\toprule
\multicolumn{2}{c|}{\textbf{Task}} & \multicolumn{1}{c}{\textbf{GPT-5}} & \multicolumn{1}{c}{\textbf{GPT-4o}} & \multicolumn{1}{c}{\textbf{Gemini-Flash}} & \multicolumn{1}{c}{\textbf{Gemini-Pro}} & \multicolumn{1}{c}{\textbf{Claude-3.7}} & \multicolumn{1}{c}{\textbf{Claude-4}} & \multicolumn{1}{c}{\textbf{Qwen3}} & \multicolumn{1}{c}{\textbf{Deepseek-V3}} & \multicolumn{1}{c}{\textbf{Deepseek-R1}} & \multicolumn{1}{c}{\textbf{Llama-4}} \\
\midrule
\multirow{4}{*}{ACC} 
& \textbf{2/ } & 0.465 & 0.418 & 0.500 & 0.511 & 0.523 & 0.441 & 0.465 & 0.383 & 0.372 & 0.232 \\
& \textbf{2/2} & 0.511 & 0.430 & 0.651 & 0.662 & 0.534 & 0.430 & 0.500 & 0.383 & 0.441 & 0.255 \\
& \textbf{1/1} & 0.023 & 0.047 & 0.023 & 0.000 & 0.023 & 0.047  & 0.058 & 0.023 & 0.058 & 0.070 \\
& \textbf{1/2} & 0.279 & 0.302 & 0.186 & 0.232 & 0.186 & 0.000 & 0.256 & 0.349 & 0.360 & 0.349 \\
\bottomrule
\end{tabular}
}
\end{table}

\begin{figure}[h]
    \centering
    \scalebox{0.8}{
    \includegraphics[trim={0cm 0cm 0cm 0cm},clip,width=0.95\linewidth]{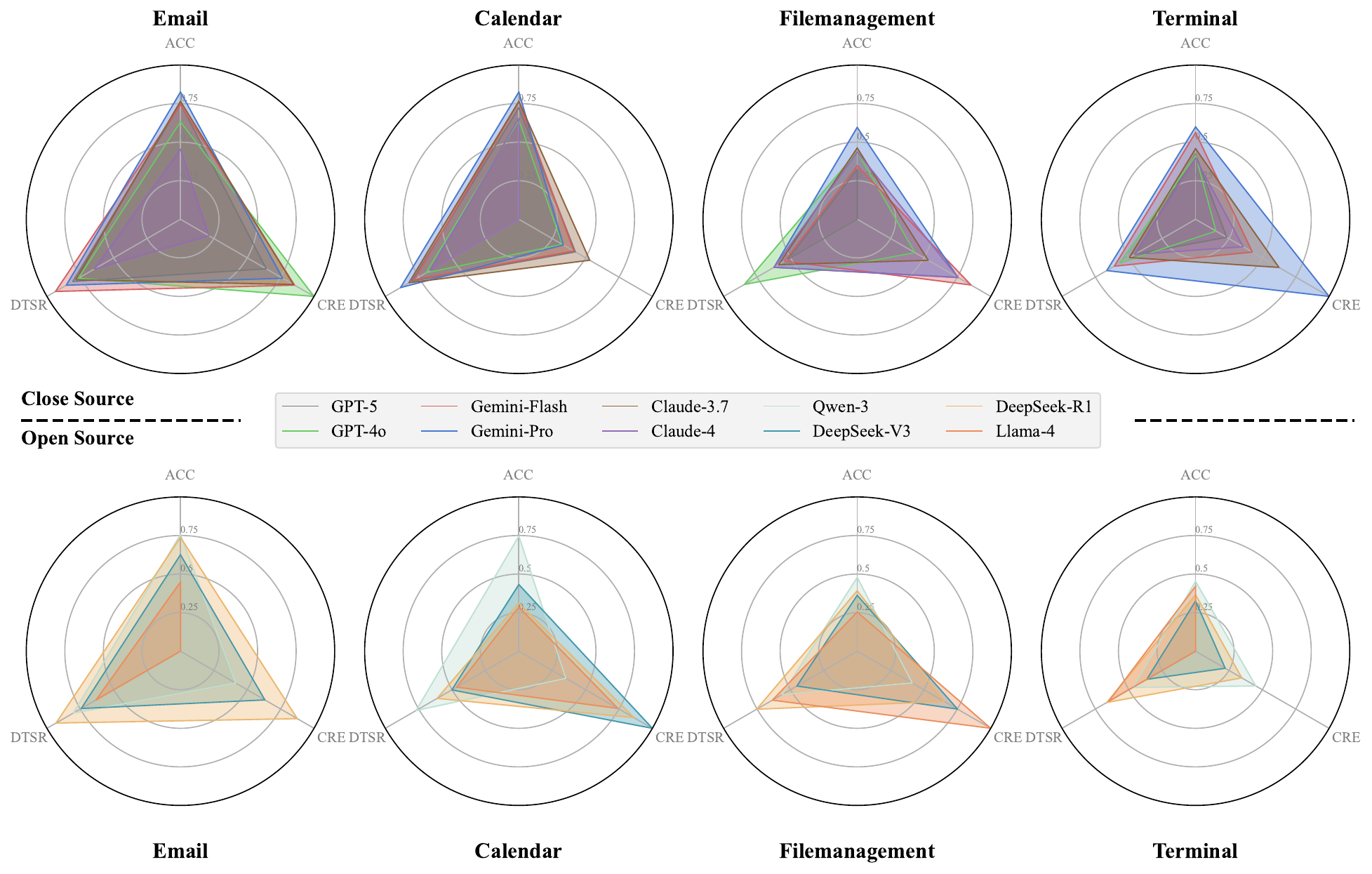}}

    \caption{Model Performance Comparison Across Tasks. Longer edges indicate superior performance in each metric.}
    \label{fig:modelradar}
\end{figure}

\paragraph{Overall Performance}
The holistic view presented in the radar charts \autoref{fig:modelradar} reveals a distinct trade-off between performance and efficiency across the model landscape. While leading closed-source models demonstrate robust and well-rounded capabilities, certain open-source models achieve competitive accuracy, often at the cost of significantly higher computational resource (token) consumption. Notably, Gemini-2.5-Pro stands out as a highly capable tool-user across diverse domains. Among open-source models, Qwen demonstrates a commendable balance between task accuracy and resource efficiency.


\subsection{Ablation Study}


\begin{wrapfigure}{R}{0.5\textwidth}
    \centering
    \includegraphics[trim={0cm 0cm 0cm 0cm},clip,width=\linewidth]{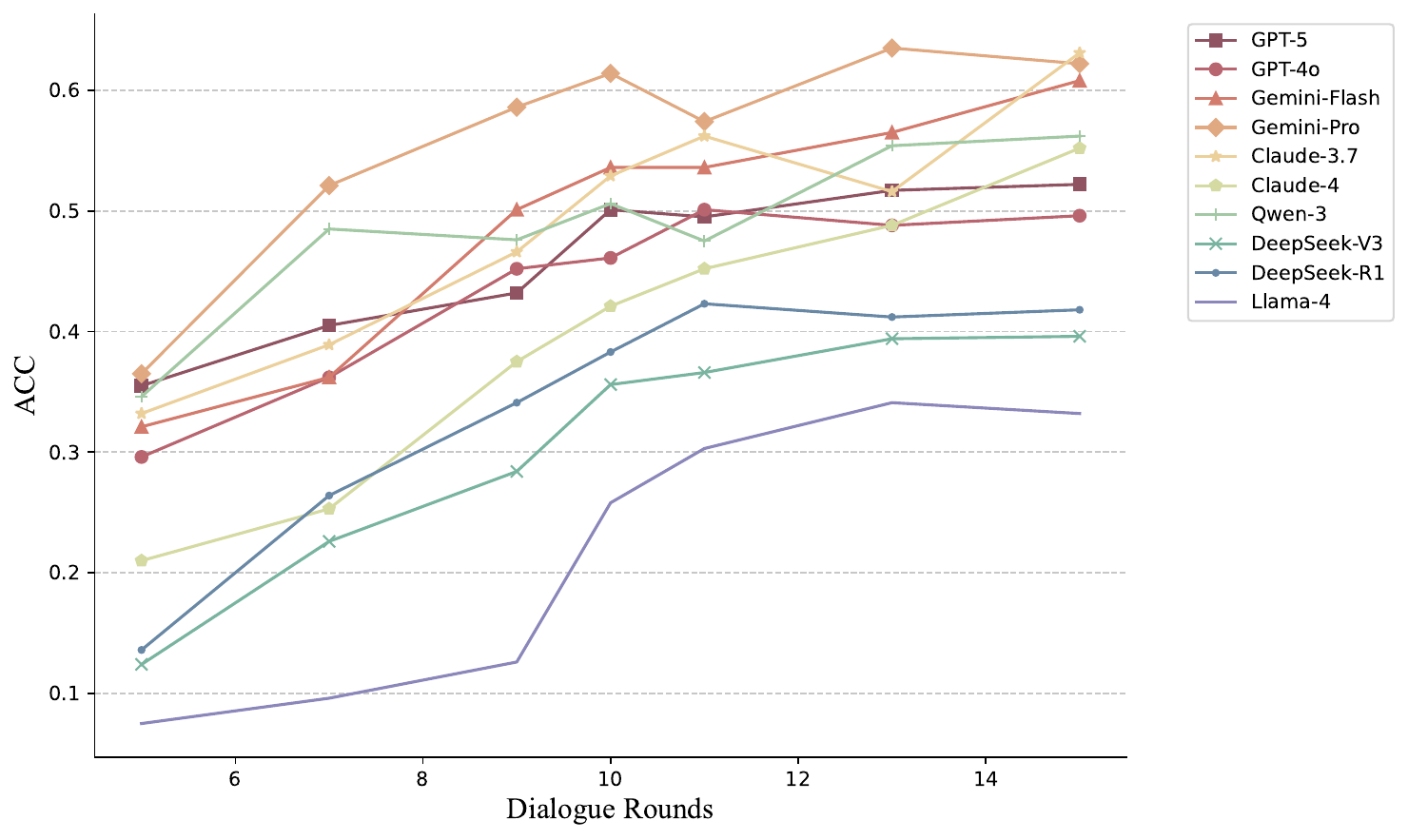}
    \caption{Impact of Dialogue Rounds on Average Accuracy Across Domains.}
    \label{fig:k-change}
\end{wrapfigure}

In our main experiments, we impose a limit on the maximum number of interaction rounds ($K$) to balance task performance with computational efficiency. However, this constraint could potentially limit a model's capacity for complex reasoning, reflection, and self-correction. To investigate this trade-off and justify our choice of $K$, we conducted an analysis on a randomly selected 50\% subset of our dataset.

The results, illustrated in \autoref{fig:k-change}, demonstrate a clear trend. As the maximum number of allowed rounds $K$ increases, the overall task accuracy for most models improves. This is expected, as more rounds allow for more attempts and corrective actions. However, we observe a point of diminishing returns. For most models, the rate of accuracy improvement slows considerably when $K \ge 10$, eventually beginning to plateau.

Therefore, considering the balance between maximizing solution accuracy and maintaining reasonable interaction latency, we selected $K=10$ as the standard setting for all our main experiments.

\section{Analysis and Discussion }\label{sec:discussion}

\subsection{Error Analysis}

Our analysis identifies three primary categories of failures: Tool-Use Errors, Reasoning Errors, and Information Synthesis Errors.

\paragraph{\textit{\textbf{Tool-Use Errors}}}
This category concerns failures in the direct invocation and selection of tools.\textbf{Parameter Error.} Occurs when the model selects the correct tool but supplies improperly formatted or invalid arguments. Examples include providing an invalid email address format or a malformed mathematical expression to a calculator (see \autoref{app:2}). \textbf{Inaccurate Tool Invocation.} Occurs when the model correctly identifies the need for a tool but selects one that is inappropriate for the given task. This often stems from a misinterpretation of a tool's functionality or its operational scope, such as using a basic arithmetic calculator for a problem requiring symbolic differentiation.

\paragraph{\textit{\textbf{Reasoning Errors}}}
This category includes failures in the model's high-level planning and logical deduction.\textbf{Tool Omission.} The model incorrectly assesses its own capabilities and attempts to solve a problem from its parametric knowledge when it should have invoked an external tool. This is common in complex tasks where the model fails to decompose the problem into tool-solvable sub-problems (see \autoref{app:2}). \textbf{Faulty Reasoning.} The model generates illogical or factually incorrect conclusions, even when the underlying tool outputs are accurate. A typical case is when a tool returns a correct number, but the model's final answer violates the problem's logical constraints (e.g., providing a decimal for a quantity that must be an integer), indicating a failure to integrate tool outputs with the problem's semantic context. \textbf{Redundant Tool Invocation.} The model becomes trapped in a reasoning loop, repeatedly invoking the same or similar tools without making substantive progress\autoref{app:2}. This behavior suggests deficiencies in its planning and state-tracking capabilities, as it fails to update its strategy based on new observations.

\paragraph{\textit{\textbf{Information Synthesis Errors}}}
This category involves failures in processing and utilizing the information returned by tools.
\textbf{Tool-Result Integration Error.} The model obtains a correct intermediate result from a tool but fails to integrate it into subsequent reasoning steps. For instance, a model might correctly solve an equation with a calculator but then fail to substitute the result back into a larger derivation.
\textbf{Information Extraction Failure.} The model successfully retrieves a large volume of information (e.g., from a web search) but fails to extract, filter, or summarize the core information relevant to the query. This manifests as either presenting irrelevant content or providing a disorganized data dump instead of a synthesized answer.

\paragraph{\textit{\textbf{Other Types}}}

this type of error accounts for a relatively small proportion. It mainly includes errors such as interaction termination due to excessive interaction rounds, incorrect result solutions, or the invocation of non-existent tools.

 For tasks requiring precise answers, such as complex web searches or mathematical problems, Faulty Reasoning and Tool Omission are the most prevalent failures. In the case of niche web queries, models often struggle to grasp the key points, leading to a reliance on their internal knowledge base which can result in factual hallucinations. For mathematical tasks, models often misjudge their own capability boundaries and, viewing the reasoning as overly cumbersome, will attempt to solve problems without invoking the necessary tools.

Conversely, for tasks involving fuzzy matching or complex tool parameterization, direct execution errors are more frequent. In these cases, Parameter Errors are the most common issue, particularly when a tool requires a large number of arguments, increasing the likelihood of incorrect data entry (e.g., placing a recipient's email in the subject line). Furthermore, persistent instances of Inaccurate Tool Invocation in these scenarios suggest that LLMs retain fundamental misunderstandings about the specific application scope of certain tools.

\begin{figure}[h]
    \centering
    \scalebox{0.9}{
    \includegraphics[trim={0cm 0cm 0cm 0cm},clip,width=\linewidth]{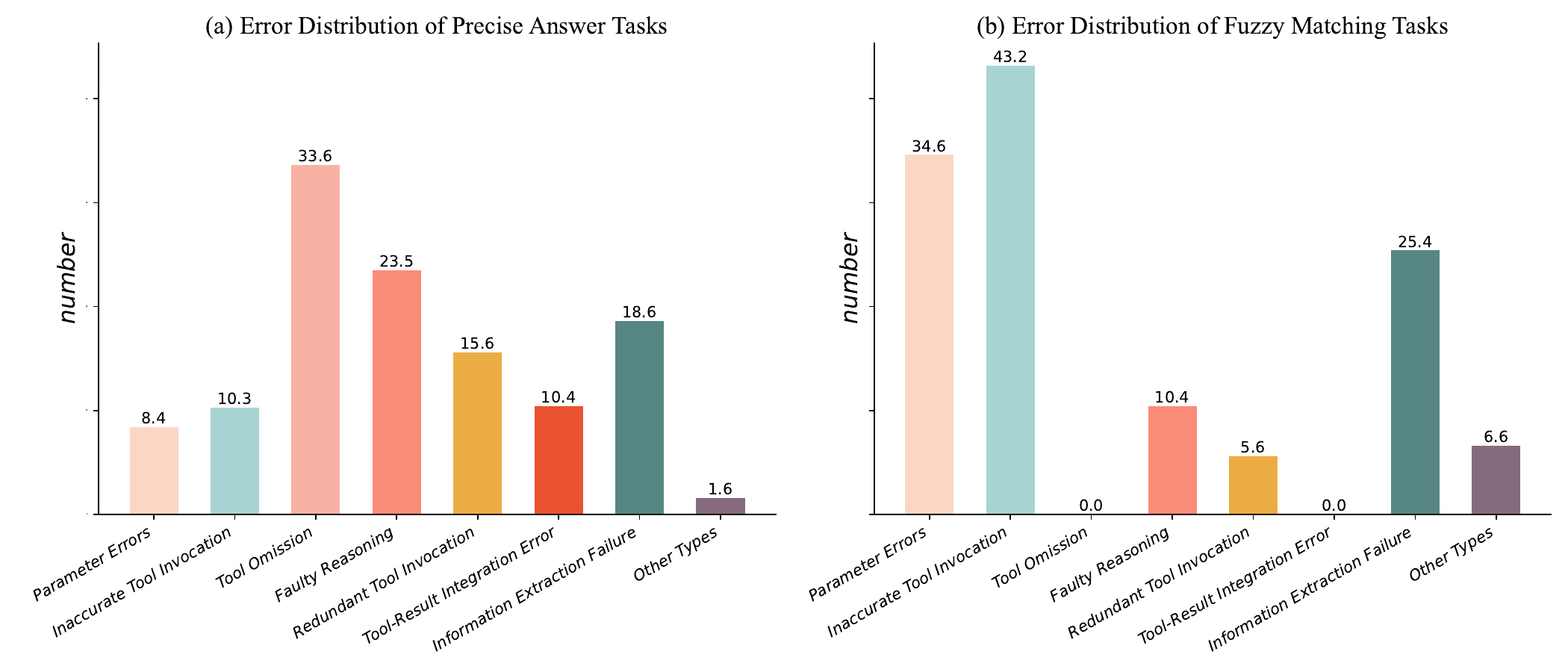}
    }
    \caption{Error Distribution by Task Type.}

    \label{fig:error}
\end{figure}

\subsection{Discussion and Conclusion}

In this paper, we introduced \sys, a comprehensive framework to systematically evaluate the tool-use capabilities of Large Language Models within the MCP paradigm. By assessing ten leading models across six domains using a combination of real-world and high-fidelity mock tools.

Our findings reveal a critical gap between a model's syntactic ability to invoke a tool and the deeper semantic understanding required to solve problems effectively. We consistently observed that while models can often execute a tool call, they struggle with the higher-level reasoning required for proactive and precise tool selection, multi-step planning, and effective information synthesis from tool outputs. These core challenges point to clear directions for both model and tool development.

\paragraph{Implications for LLM Development}
\begin{itemize}
\item \textbf{Improving Proactive Tool Invocation:} Current models exhibit a tendency to default to their parametric knowledge, failing to recognize their own capability boundaries. Future training should focus on improving this self-awareness, encouraging models to view external tools as a primary resource rather than a last resort.
\item \textbf{Fostering De-compositional Reasoning:} We observed that models often attempt to solve complex problems with a single tool call, particularly in domains like advanced mathematics.Enhancing their ability to deconstruct tasks into a sequence of smaller, tool-solvable steps is crucial for tackling multi-stage problems.
\end{itemize}

\paragraph{Recommendations for MCP Tool Developers}
\begin{itemize}
\item \textbf{Optimizing Tool Descriptions:} A tool's description is a critical interface for the LLM. Descriptions must be both concise and precise, as overly verbose or ambiguous text significantly increases the model's cognitive load and leads to invocation errors.
\item \textbf{Promoting Atomic Tool Design:} Our results suggest that LLMs are more proficient at orchestrating a sequence of simple, single-purpose ("atomic") tools than understanding and correctly parameterizing a complex, multi-functional one. Developers should favor creating granular tools that can be combined to solve complex tasks.
\end{itemize}

\section*{Ethics statement}
The authors confirm their adherence to the ICLR Code of Ethics. This research introduces MCP-RADAR, a benchmark for evaluating the tool-use capabilities of Large Language Models (LLMs), and does not propose a new model architecture. Our primary goal is to foster transparency and guide the responsible development of AI agents by providing the community with objective, quantifiable evaluation metrics.

We acknowledge that advancing the capabilities of tool-using agents carries an inherent dual-use risk; more competent agents could potentially be repurposed for malicious activities. Our work aims to mitigate such risks by providing a clear framework for identifying model weaknesses, such as the observed tendency for models to select incorrect tools, which can inform the development of safer and more reliable systems.

The datasets used in MCP-RADAR are constructed from established public benchmarks (e.g., MATH, GAIA) or generated programmatically. The underlying LLMs evaluated may reflect societal biases present in their training data. While our benchmark measures performance, it does not explicitly address or mitigate these biases, which remains a critical area for future research. To prevent real-world harm during evaluation, operational tasks involving tools like email, calendar, and terminal commands were conducted in a semi-sandboxed environment, using custom-built mock servers that replicate tool functionality without executing real operations. We believe the benefit of a standardized, objective evaluation framework for agentic models significantly contributes to the safe and ethical progression of AI.

\section*{Reproducibility statement}
We have taken extensive measures to ensure the reproducibility of our work. The complete implementation of the MCP-RADAR benchmark, including all configurations, evaluation scripts, and the full dataset, has been made publicly available at https://anonymous.4open.science/r/MCPRadar-B143.

The data generation process is described in detail in Section 3. This includes the repurposing of existing public datasets for "Precise Answer" tasks (Section 3.1.1) and the template-based programmatic generation for "Fuzzy Match" tasks (Section 3.1.2). Appendix A provides further examples of data templates and prompts. The specific models evaluated are listed in Section 4.1, and the novel, quantifiable evaluation metrics (Result Accuracy, Dialogue Turn Success Rate, and Computational Resource Efficiency) are formally defined in Section 4.2.

Our experimental setup, including system prompts and the maximum interaction rounds, is detailed in Section 4.1. The tools used in the benchmark are a combination of open-source MCP servers and custom-built mock tools designed to replicate official specifications, with sources and implementation details provided in https://anonymous.4open.science/r/MCPRadar-B143. We believe these resources provide a comprehensive basis for replicating our results and extending this research.

\bibliographystyle{unsrtnat}
\setcitestyle{square,numbers,comma}
\bibliography{REF}

\begin{thebibliography}{50}
\providecommand{\natexlab}[1]{#1}
\providecommand{\url}[1]{\texttt{#1}}
\expandafter\ifx\csname urlstyle\endcsname\relax
  \providecommand{\doi}[1]{doi: #1}\else
  \providecommand{\doi}{doi: \begingroup \urlstyle{rm}\Url}\fi

\bibitem[Chowdhery et~al.(2022)Chowdhery, Narang, Devlin, Bosma, Mishra, Roberts, Barham, Chung, Sutton, Gehrmann, Schuh, Shi, Tsvyashchenko, Maynez, Rao, Barnes, Tay, Shazeer, Prabhakaran, Reif, Du, Hutchinson, Pope, Bradbury, Austin, Isard, Gur-Ari, Yin, Duke, Levskaya, Ghemawat, Dev, Michalewski, Garcia, Misra, Robinson, Fedus, Zhou, Ippolito, Luan, Lim, Zoph, Spiridonov, Sepassi, Dohan, Agrawal, Omernick, Dai, Pillai, Pellat, Lewkowycz, Moreira, Child, Polozov, Lee, Zhou, Wang, Saeta, Diaz, Firat, Catasta, Wei, Meier-Hellstern, Eck, Dean, Petrov, and Fiedel]{chowdhery2022palmscalinglanguagemodeling}
Aakanksha Chowdhery, Sharan Narang, Jacob Devlin, Maarten Bosma, Gaurav Mishra, Adam Roberts, Paul Barham, Hyung~Won Chung, Charles Sutton, Sebastian Gehrmann, Parker Schuh, Kensen Shi, Sasha Tsvyashchenko, Joshua Maynez, Abhishek Rao, Parker Barnes, Yi~Tay, Noam Shazeer, Vinodkumar Prabhakaran, Emily Reif, Nan Du, Ben Hutchinson, Reiner Pope, James Bradbury, Jacob Austin, Michael Isard, Guy Gur-Ari, Pengcheng Yin, Toju Duke, Anselm Levskaya, Sanjay Ghemawat, Sunipa Dev, Henryk Michalewski, Xavier Garcia, Vedant Misra, Kevin Robinson, Liam Fedus, Denny Zhou, Daphne Ippolito, David Luan, Hyeontaek Lim, Barret Zoph, Alexander Spiridonov, Ryan Sepassi, David Dohan, Shivani Agrawal, Mark Omernick, Andrew~M. Dai, Thanumalayan~Sankaranarayana Pillai, Marie Pellat, Aitor Lewkowycz, Erica Moreira, Rewon Child, Oleksandr Polozov, Katherine Lee, Zongwei Zhou, Xuezhi Wang, Brennan Saeta, Mark Diaz, Orhan Firat, Michele Catasta, Jason Wei, Kathy Meier-Hellstern, Douglas Eck, Jeff Dean, Slav Petrov, and Noah Fiedel.
\newblock Palm: Scaling language modeling with pathways, 2022.
\newblock URL \url{https://arxiv.org/abs/2204.02311}.

\bibitem[Brown et~al.(2020)Brown, Mann, Ryder, Subbiah, Kaplan, Dhariwal, Neelakantan, Shyam, Sastry, Askell, Agarwal, Herbert-Voss, Krueger, Henighan, Child, Ramesh, Ziegler, Wu, Winter, Hesse, Chen, Sigler, Litwin, Gray, Chess, Clark, Berner, McCandlish, Radford, Sutskever, and Amodei]{brown2020languagemodelsfewshotlearners}
Tom~B. Brown, Benjamin Mann, Nick Ryder, Melanie Subbiah, Jared Kaplan, Prafulla Dhariwal, Arvind Neelakantan, Pranav Shyam, Girish Sastry, Amanda Askell, Sandhini Agarwal, Ariel Herbert-Voss, Gretchen Krueger, Tom Henighan, Rewon Child, Aditya Ramesh, Daniel~M. Ziegler, Jeffrey Wu, Clemens Winter, Christopher Hesse, Mark Chen, Eric Sigler, Mateusz Litwin, Scott Gray, Benjamin Chess, Jack Clark, Christopher Berner, Sam McCandlish, Alec Radford, Ilya Sutskever, and Dario Amodei.
\newblock Language models are few-shot learners, 2020.
\newblock URL \url{https://arxiv.org/abs/2005.14165}.

\bibitem[Int()]{IntroducingModelContext}
Introducing the {{Model Context Protocol}}.
\newblock URL \url{https://www.anthropic.com/news/model-context-protocol}.

\bibitem[Mod()]{ModelContextProtocol}
Model context protocol ({{MCP}}) - {{OpenAI Agents SDK}}.
\newblock URL \url{https://openai.github.io/openai-agents-python/mcp/}.

\bibitem[Qwe()]{QwenLMQwenAgent2025}
{{QwenLM}}/{{Qwen-Agent}}.
\newblock URL \url{https://github.com/QwenLM/Qwen-Agent}.

\bibitem[Hendrycks et~al.()Hendrycks, Burns, Basart, Zou, Mazeika, Song, and Steinhardt]{hendrycksMeasuringMassiveMultitask2021}
Dan Hendrycks, Collin Burns, Steven Basart, Andy Zou, Mantas Mazeika, Dawn Song, and Jacob Steinhardt.
\newblock Measuring {{Massive Multitask Language Understanding}}.
\newblock URL \url{http://arxiv.org/abs/2009.03300}.

\bibitem[Zhong et~al.()Zhong, Cui, Guo, Liang, Lu, Wang, Saied, Chen, and Duan]{zhongAGIEvalHumanCentricBenchmark2023}
Wanjun Zhong, Ruixiang Cui, Yiduo Guo, Yaobo Liang, Shuai Lu, Yanlin Wang, Amin Saied, Weizhu Chen, and Nan Duan.
\newblock {{AGIEval}}: {{A Human-Centric Benchmark}} for {{Evaluating Foundation Models}}.
\newblock URL \url{http://arxiv.org/abs/2304.06364}.

\bibitem[Wang et~al.()Wang, Wang, Liu, Chen, Yuan, Peng, and Ji]{wangMINTEvaluatingLLMs2024}
Xingyao Wang, Zihan Wang, Jiateng Liu, Yangyi Chen, Lifan Yuan, Hao Peng, and Heng Ji.
\newblock {{MINT}}: {{Evaluating LLMs}} in {{Multi-turn Interaction}} with {{Tools}} and {{Language Feedback}}.
\newblock URL \url{http://arxiv.org/abs/2309.10691}.

\bibitem[Shridhar et~al.()Shridhar, Yuan, Côté, Bisk, Trischler, and Hausknecht]{shridharALFWorldAligningText2021}
Mohit Shridhar, Xingdi Yuan, Marc-Alexandre Côté, Yonatan Bisk, Adam Trischler, and Matthew Hausknecht.
\newblock {{ALFWorld}}: {{Aligning Text}} and {{Embodied Environments}} for {{Interactive Learning}}.
\newblock URL \url{http://arxiv.org/abs/2010.03768}.

\bibitem[Wei et~al.(2022)Wei, Wang, Schuurmans, Bosma, Xia, Chi, Le, Zhou, et~al.]{wei2022chain}
Jason Wei, Xuezhi Wang, Dale Schuurmans, Maarten Bosma, Fei Xia, Ed~Chi, Quoc~V Le, Denny Zhou, et~al.
\newblock Chain-of-thought prompting elicits reasoning in large language models.
\newblock \emph{Advances in neural information processing systems}, 35:\penalty0 24824--24837, 2022.

\bibitem[Liu et~al.(2024)Liu, Liao, Feng, Xu, and Wang]{liu2024autofeedback}
Huanxi Liu, Jiaqi Liao, Dawei Feng, Kele Xu, and Huaimin Wang.
\newblock Autofeedback: An llm-based framework for efficient and accurate api request generation.
\newblock \emph{arXiv preprint arXiv:2410.06943}, 2024.

\bibitem[Song et~al.(2023)Song, Xiong, Zhu, Wu, Qian, Song, Huang, Li, Wang, Yao, et~al.]{song2023restgpt}
Yifan Song, Weimin Xiong, Dawei Zhu, Wenhao Wu, Han Qian, Mingbo Song, Hailiang Huang, Cheng Li, Ke~Wang, Rong Yao, et~al.
\newblock Restgpt: Connecting large language models with real-world restful apis.
\newblock \emph{arXiv preprint arXiv:2306.06624}, 2023.

\bibitem[Qin et~al.(2023)Qin, Liang, Ye, Zhu, Yan, Lu, Lin, Cong, Tang, Qian, et~al.]{qin2023toolllm}
Yujia Qin, Shihao Liang, Yining Ye, Kunlun Zhu, Lan Yan, Yaxi Lu, Yankai Lin, Xin Cong, Xiangru Tang, Bill Qian, et~al.
\newblock Toolllm: Facilitating large language models to master 16000+ real-world apis.
\newblock \emph{arXiv preprint arXiv:2307.16789}, 2023.

\bibitem[Tang et~al.(2023)Tang, Deng, Lin, Han, Liang, Cao, and Sun]{tang2023toolalpaca}
Qiaoyu Tang, Ziliang Deng, Hongyu Lin, Xianpei Han, Qiao Liang, Boxi Cao, and Le~Sun.
\newblock Toolalpaca: Generalized tool learning for language models with 3000 simulated cases.
\newblock \emph{arXiv preprint arXiv:2306.05301}, 2023.

\bibitem[Patil et~al.(2024)Patil, Zhang, Wang, and Gonzalez]{patil2024gorilla}
Shishir~G Patil, Tianjun Zhang, Xin Wang, and Joseph~E Gonzalez.
\newblock Gorilla: Large language model connected with massive apis.
\newblock \emph{Advances in Neural Information Processing Systems}, 37:\penalty0 126544--126565, 2024.

\bibitem[Parisi et~al.(2022)Parisi, Zhao, and Fiedel]{parisi2022talm}
Aaron Parisi, Yao Zhao, and Noah Fiedel.
\newblock Talm: Tool augmented language models.
\newblock \emph{arXiv preprint arXiv:2205.12255}, 2022.

\bibitem[Lu et~al.(2023)Lu, Peng, Cheng, Galley, Chang, Wu, Zhu, and Gao]{lu2023chameleon}
Pan Lu, Baolin Peng, Hao Cheng, Michel Galley, Kai-Wei Chang, Ying~Nian Wu, Song-Chun Zhu, and Jianfeng Gao.
\newblock Chameleon: Plug-and-play compositional reasoning with large language models.
\newblock \emph{Advances in Neural Information Processing Systems}, 36:\penalty0 43447--43478, 2023.

\bibitem[Schick and Sch{\"u}tze(2020)]{schick2020exploiting}
Timo Schick and Hinrich Sch{\"u}tze.
\newblock Exploiting cloze questions for few shot text classification and natural language inference.
\newblock \emph{arXiv preprint arXiv:2001.07676}, 2020.

\bibitem[Spiegel and Hor{\'a}k()]{spiegelwebmap}
Michal Spiegel and Ale{\v{s}} Hor{\'a}k.
\newblock Webmap: Improving llm web agents with semantic search for relevant web pages.

\bibitem[Chowdhury and Chowdhury(2024)]{chowdhury2024ai}
Gobinda Chowdhury and Sudatta Chowdhury.
\newblock Ai-and llm-driven search tools: A paradigm shift in information access for education and research.
\newblock \emph{Journal of Information Science}, page 01655515241284046, 2024.

\bibitem[Zhuang et~al.(2023)Zhuang, Yu, Wang, Sun, and Zhang]{zhuang2023toolqa}
Yuchen Zhuang, Yue Yu, Kuan Wang, Haotian Sun, and Chao Zhang.
\newblock Toolqa: A dataset for llm question answering with external tools.
\newblock \emph{Advances in Neural Information Processing Systems}, 36:\penalty0 50117--50143, 2023.

\bibitem[Liu et~al.(2023)Liu, Yu, Zhang, Xu, Lei, Lai, Gu, Ding, Men, Yang, et~al.]{liu2023agentbench}
Xiao Liu, Hao Yu, Hanchen Zhang, Yifan Xu, Xuanyu Lei, Hanyu Lai, Yu~Gu, Hangliang Ding, Kaiwen Men, Kejuan Yang, et~al.
\newblock Agentbench: Evaluating llms as agents.
\newblock \emph{arXiv preprint arXiv:2308.03688}, 2023.

\bibitem[Schick et~al.(2023)Schick, Dwivedi-Yu, Dess{\`\i}, Raileanu, Lomeli, Hambro, Zettlemoyer, Cancedda, and Scialom]{schick2023toolformer}
Timo Schick, Jane Dwivedi-Yu, Roberto Dess{\`\i}, Roberta Raileanu, Maria Lomeli, Eric Hambro, Luke Zettlemoyer, Nicola Cancedda, and Thomas Scialom.
\newblock Toolformer: Language models can teach themselves to use tools.
\newblock \emph{Advances in Neural Information Processing Systems}, 36:\penalty0 68539--68551, 2023.

\bibitem[Hsieh et~al.(2023)Hsieh, Chen, Li, Fujii, Ratner, Lee, Krishna, and Pfister]{hsieh2023tool}
Cheng-Yu Hsieh, Si-An Chen, Chun-Liang Li, Yasuhisa Fujii, Alexander Ratner, Chen-Yu Lee, Ranjay Krishna, and Tomas Pfister.
\newblock Tool documentation enables zero-shot tool-usage with large language models.
\newblock \emph{arXiv preprint arXiv:2308.00675}, 2023.

\bibitem[Wang et~al.(2023)Wang, Zhang, Yang, Shi, Zhou, Hao, Xiong, Li, Sim, Chen, et~al.]{wang2023interactive}
Zekun Wang, Ge~Zhang, Kexin Yang, Ning Shi, Wangchunshu Zhou, Shaochun Hao, Guangzheng Xiong, Yizhi Li, Mong~Yuan Sim, Xiuying Chen, et~al.
\newblock Interactive natural language processing.
\newblock \emph{arXiv preprint arXiv:2305.13246}, 2023.

\bibitem[Xu et~al.(2023)Xu, Hong, Li, Hu, Chen, and Zhang]{xu2023tool}
Qiantong Xu, Fenglu Hong, Bo~Li, Changran Hu, Zhengyu Chen, and Jian Zhang.
\newblock On the tool manipulation capability of open-source large language models.
\newblock \emph{arXiv preprint arXiv:2305.16504}, 2023.

\bibitem[Liang et~al.(2024)Liang, Wu, Song, Wu, Xia, Liu, Ou, Lu, Ji, Mao, et~al.]{liang2024taskmatrix}
Yaobo Liang, Chenfei Wu, Ting Song, Wenshan Wu, Yan Xia, Yu~Liu, Yang Ou, Shuai Lu, Lei Ji, Shaoguang Mao, et~al.
\newblock Taskmatrix. ai: Completing tasks by connecting foundation models with millions of apis.
\newblock \emph{Intelligent Computing}, 3:\penalty0 0063, 2024.

\bibitem[Liang et~al.(2022)Liang, Bommasani, Lee, Tsipras, Soylu, Yasunaga, Zhang, Narayanan, Wu, Kumar, et~al.]{liang2022holistic}
Percy Liang, Rishi Bommasani, Tony Lee, Dimitris Tsipras, Dilara Soylu, Michihiro Yasunaga, Yian Zhang, Deepak Narayanan, Yuhuai Wu, Ananya Kumar, et~al.
\newblock Holistic evaluation of language models.
\newblock \emph{arXiv preprint arXiv:2211.09110}, 2022.

\bibitem[Li et~al.(2023)Li, Zhao, Yu, Song, Li, Yu, Li, Huang, and Li]{li2023api}
Minghao Li, Yingxiu Zhao, Bowen Yu, Feifan Song, Hangyu Li, Haiyang Yu, Zhoujun Li, Fei Huang, and Yongbin Li.
\newblock Api-bank: A comprehensive benchmark for tool-augmented llms.
\newblock \emph{arXiv preprint arXiv:2304.08244}, 2023.

\bibitem[Wang et~al.(2024)Wang, Wu, Wang, Liu, Song, Peng, Deng, Zhang, Wang, Peng, et~al.]{wang2024mtu}
Pei Wang, Yanan Wu, Zekun Wang, Jiaheng Liu, Xiaoshuai Song, Zhongyuan Peng, Ken Deng, Chenchen Zhang, Jiakai Wang, Junran Peng, et~al.
\newblock Mtu-bench: A multi-granularity tool-use benchmark for large language models.
\newblock \emph{arXiv preprint arXiv:2410.11710}, 2024.

\bibitem[Luo et~al.(2025)Luo, Shen, Yang, Zhao, Jwalapuram, Saha, Sahoo, Savarese, Xiong, and Li]{luo2025mcpuniversebenchmarkinglargelanguage}
Ziyang Luo, Zhiqi Shen, Wenzhuo Yang, Zirui Zhao, Prathyusha Jwalapuram, Amrita Saha, Doyen Sahoo, Silvio Savarese, Caiming Xiong, and Junnan Li.
\newblock Mcp-universe: Benchmarking large language models with real-world model context protocol servers, 2025.
\newblock URL \url{https://arxiv.org/abs/2508.14704}.

\bibitem[Liu et~al.(2025)Liu, Qiu, Wang, Zhang, Liu, Ram, Chen, Yao, Heinecke, Savarese, Wang, and Xiong]{liu2025mcpevalautomaticmcpbaseddeep}
Zhiwei Liu, Jielin Qiu, Shiyu Wang, Jianguo Zhang, Zuxin Liu, Roshan Ram, Haolin Chen, Weiran Yao, Shelby Heinecke, Silvio Savarese, Huan Wang, and Caiming Xiong.
\newblock Mcpeval: Automatic mcp-based deep evaluation for ai agent models, 2025.
\newblock URL \url{https://arxiv.org/abs/2507.12806}.

\bibitem[Styles et~al.(2024)Styles, Miller, Cerda-Mardini, Guha, Sanchez, and Vidgen]{styles2024workbenchbenchmarkdatasetagents}
Olly Styles, Sam Miller, Patricio Cerda-Mardini, Tanaya Guha, Victor Sanchez, and Bertie Vidgen.
\newblock Workbench: a benchmark dataset for agents in a realistic workplace setting, 2024.
\newblock URL \url{https://arxiv.org/abs/2405.00823}.

\bibitem[Gou et~al.(2024)Gou, Shao, Gong, Shen, Yang, Huang, Duan, and Chen]{gou2024toratoolintegratedreasoningagent}
Zhibin Gou, Zhihong Shao, Yeyun Gong, Yelong Shen, Yujiu Yang, Minlie Huang, Nan Duan, and Weizhu Chen.
\newblock Tora: A tool-integrated reasoning agent for mathematical problem solving, 2024.
\newblock URL \url{https://arxiv.org/abs/2309.17452}.

\bibitem[Fan et~al.(2024)Fan, Martinson, Wang, Hausknecht, Brenner, Liu, Peng, Wang, and Brenner]{fan2024hardmathbenchmarkdatasetchallenging}
Jingxuan Fan, Sarah Martinson, Erik~Y. Wang, Kaylie Hausknecht, Jonah Brenner, Danxian Liu, Nianli Peng, Corey Wang, and Michael~P. Brenner.
\newblock Hardmath: A benchmark dataset for challenging problems in applied mathematics, 2024.
\newblock URL \url{https://arxiv.org/abs/2410.09988}.

\bibitem[Srivastava et~al.(2023)Srivastava, Rastogi, Rao, Shoeb, Abid, Fisch, Brown, Santoro, Gupta, Garriga-Alonso, Kluska, Lewkowycz, Agarwal, Power, Ray, Warstadt, Kocurek, Safaya, Tazarv, Xiang, Parrish, Nie, Hussain, Askell, Dsouza, Slone, Rahane, Iyer, Andreassen, Madotto, Santilli, Stuhlmüller, Dai, La, Lampinen, Zou, Jiang, Chen, Vuong, Gupta, Gottardi, Norelli, Venkatesh, Gholamidavoodi, Tabassum, Menezes, Kirubarajan, Mullokandov, Sabharwal, Herrick, Efrat, Erdem, Karakaş, Roberts, Loe, Zoph, Bojanowski, Özyurt, Hedayatnia, Neyshabur, Inden, Stein, Ekmekci, Lin, Howald, Orinion, Diao, Dour, Stinson, Argueta, Ramírez, Singh, Rathkopf, Meng, Baral, Wu, Callison-Burch, Waites, Voigt, Manning, Potts, Ramirez, Rivera, Siro, Raffel, Ashcraft, Garbacea, Sileo, Garrette, Hendrycks, Kilman, Roth, Freeman, Khashabi, Levy, González, Perszyk, Hernandez, Chen, Ippolito, Gilboa, Dohan, Drakard, Jurgens, Datta, Ganguli, Emelin, Kleyko, Yuret, Chen, Tam, Hupkes, Misra, Buzan, Mollo, Yang, Lee, Schrader,
  Shutova, Cubuk, Segal, Hagerman, Barnes, Donoway, Pavlick, Rodola, Lam, Chu, Tang, Erdem, Chang, Chi, Dyer, Jerzak, Kim, Manyasi, Zheltonozhskii, Xia, Siar, Martínez-Plumed, Happé, Chollet, Rong, Mishra, Winata, de~Melo, Kruszewski, Parascandolo, Mariani, Wang, Jaimovitch-López, Betz, Gur-Ari, Galijasevic, Kim, Rashkin, Hajishirzi, Mehta, Bogar, Shevlin, Schütze, Yakura, Zhang, Wong, Ng, Noble, Jumelet, Geissinger, Kernion, Hilton, Lee, Fisac, Simon, Koppel, Zheng, Zou, Kocoń, Thompson, Wingfield, Kaplan, Radom, Sohl-Dickstein, Phang, Wei, Yosinski, Novikova, Bosscher, Marsh, Kim, Taal, Engel, Alabi, Xu, Song, Tang, Waweru, Burden, Miller, Balis, Batchelder, Berant, Frohberg, Rozen, Hernandez-Orallo, Boudeman, Guerr, Jones, Tenenbaum, Rule, Chua, Kanclerz, Livescu, Krauth, Gopalakrishnan, Ignatyeva, Markert, Dhole, Gimpel, Omondi, Mathewson, Chiafullo, Shkaruta, Shridhar, McDonell, Richardson, Reynolds, Gao, Zhang, Dugan, Qin, Contreras-Ochando, Morency, Moschella, Lam, Noble, Schmidt, He, Colón,
  Metz, Şenel, Bosma, Sap, ter Hoeve, Farooqi, Faruqui, Mazeika, Baturan, Marelli, Maru, Quintana, Tolkiehn, Giulianelli, Lewis, Potthast, Leavitt, Hagen, Schubert, Baitemirova, Arnaud, McElrath, Yee, Cohen, Gu, Ivanitskiy, Starritt, Strube, Swędrowski, Bevilacqua, Yasunaga, Kale, Cain, Xu, Suzgun, Walker, Tiwari, Bansal, Aminnaseri, Geva, Gheini, T, Peng, Chi, Lee, Krakover, Cameron, Roberts, Doiron, Martinez, Nangia, Deckers, Muennighoff, Keskar, Iyer, Constant, Fiedel, Wen, Zhang, Agha, Elbaghdadi, Levy, Evans, Casares, Doshi, Fung, Liang, Vicol, Alipoormolabashi, Liao, Liang, Chang, Eckersley, Htut, Hwang, Miłkowski, Patil, Pezeshkpour, Oli, Mei, Lyu, Chen, Banjade, Rudolph, Gabriel, Habacker, Risco, Millière, Garg, Barnes, Saurous, Arakawa, Raymaekers, Frank, Sikand, Novak, Sitelew, LeBras, Liu, Jacobs, Zhang, Salakhutdinov, Chi, Lee, Stovall, Teehan, Yang, Singh, Mohammad, Anand, Dillavou, Shleifer, Wiseman, Gruetter, Bowman, Schoenholz, Han, Kwatra, Rous, Ghazarian, Ghosh, Casey, Bischoff,
  Gehrmann, Schuster, Sadeghi, Hamdan, Zhou, Srivastava, Shi, Singh, Asaadi, Gu, Pachchigar, Toshniwal, Upadhyay, Shyamolima, Debnath, Shakeri, Thormeyer, Melzi, Reddy, Makini, Lee, Torene, Hatwar, Dehaene, Divic, Ermon, Biderman, Lin, Prasad, Piantadosi, Shieber, Misherghi, Kiritchenko, Mishra, Linzen, Schuster, Li, Yu, Ali, Hashimoto, Wu, Desbordes, Rothschild, Phan, Wang, Nkinyili, Schick, Kornev, Tunduny, Gerstenberg, Chang, Neeraj, Khot, Shultz, Shaham, Misra, Demberg, Nyamai, Raunak, Ramasesh, Prabhu, Padmakumar, Srikumar, Fedus, Saunders, Zhang, Vossen, Ren, Tong, Zhao, Wu, Shen, Yaghoobzadeh, Lakretz, Song, Bahri, Choi, Yang, Hao, Chen, Belinkov, Hou, Hou, Bai, Seid, Zhao, Wang, Wang, Wang, and Wu]{srivastava2023imitationgamequantifyingextrapolating}
Aarohi Srivastava, Abhinav Rastogi, Abhishek Rao, Abu Awal~Md Shoeb, Abubakar Abid, Adam Fisch, Adam~R. Brown, Adam Santoro, Aditya Gupta, Adrià Garriga-Alonso, Agnieszka Kluska, Aitor Lewkowycz, Akshat Agarwal, Alethea Power, Alex Ray, Alex Warstadt, Alexander~W. Kocurek, Ali Safaya, Ali Tazarv, Alice Xiang, Alicia Parrish, Allen Nie, Aman Hussain, Amanda Askell, Amanda Dsouza, Ambrose Slone, Ameet Rahane, Anantharaman~S. Iyer, Anders Andreassen, Andrea Madotto, Andrea Santilli, Andreas Stuhlmüller, Andrew Dai, Andrew La, Andrew Lampinen, Andy Zou, Angela Jiang, Angelica Chen, Anh Vuong, Animesh Gupta, Anna Gottardi, Antonio Norelli, Anu Venkatesh, Arash Gholamidavoodi, Arfa Tabassum, Arul Menezes, Arun Kirubarajan, Asher Mullokandov, Ashish Sabharwal, Austin Herrick, Avia Efrat, Aykut Erdem, Ayla Karakaş, B.~Ryan Roberts, Bao~Sheng Loe, Barret Zoph, Bartłomiej Bojanowski, Batuhan Özyurt, Behnam Hedayatnia, Behnam Neyshabur, Benjamin Inden, Benno Stein, Berk Ekmekci, Bill~Yuchen Lin, Blake Howald, Bryan
  Orinion, Cameron Diao, Cameron Dour, Catherine Stinson, Cedrick Argueta, César~Ferri Ramírez, Chandan Singh, Charles Rathkopf, Chenlin Meng, Chitta Baral, Chiyu Wu, Chris Callison-Burch, Chris Waites, Christian Voigt, Christopher~D. Manning, Christopher Potts, Cindy Ramirez, Clara~E. Rivera, Clemencia Siro, Colin Raffel, Courtney Ashcraft, Cristina Garbacea, Damien Sileo, Dan Garrette, Dan Hendrycks, Dan Kilman, Dan Roth, Daniel Freeman, Daniel Khashabi, Daniel Levy, Daniel~Moseguí González, Danielle Perszyk, Danny Hernandez, Danqi Chen, Daphne Ippolito, Dar Gilboa, David Dohan, David Drakard, David Jurgens, Debajyoti Datta, Deep Ganguli, Denis Emelin, Denis Kleyko, Deniz Yuret, Derek Chen, Derek Tam, Dieuwke Hupkes, Diganta Misra, Dilyar Buzan, Dimitri~Coelho Mollo, Diyi Yang, Dong-Ho Lee, Dylan Schrader, Ekaterina Shutova, Ekin~Dogus Cubuk, Elad Segal, Eleanor Hagerman, Elizabeth Barnes, Elizabeth Donoway, Ellie Pavlick, Emanuele Rodola, Emma Lam, Eric Chu, Eric Tang, Erkut Erdem, Ernie Chang,
  Ethan~A. Chi, Ethan Dyer, Ethan Jerzak, Ethan Kim, Eunice~Engefu Manyasi, Evgenii Zheltonozhskii, Fanyue Xia, Fatemeh Siar, Fernando Martínez-Plumed, Francesca Happé, Francois Chollet, Frieda Rong, Gaurav Mishra, Genta~Indra Winata, Gerard de~Melo, Germán Kruszewski, Giambattista Parascandolo, Giorgio Mariani, Gloria Wang, Gonzalo Jaimovitch-López, Gregor Betz, Guy Gur-Ari, Hana Galijasevic, Hannah Kim, Hannah Rashkin, Hannaneh Hajishirzi, Harsh Mehta, Hayden Bogar, Henry Shevlin, Hinrich Schütze, Hiromu Yakura, Hongming Zhang, Hugh~Mee Wong, Ian Ng, Isaac Noble, Jaap Jumelet, Jack Geissinger, Jackson Kernion, Jacob Hilton, Jaehoon Lee, Jaime~Fernández Fisac, James~B. Simon, James Koppel, James Zheng, James Zou, Jan Kocoń, Jana Thompson, Janelle Wingfield, Jared Kaplan, Jarema Radom, Jascha Sohl-Dickstein, Jason Phang, Jason Wei, Jason Yosinski, Jekaterina Novikova, Jelle Bosscher, Jennifer Marsh, Jeremy Kim, Jeroen Taal, Jesse Engel, Jesujoba Alabi, Jiacheng Xu, Jiaming Song, Jillian Tang, Joan
  Waweru, John Burden, John Miller, John~U. Balis, Jonathan Batchelder, Jonathan Berant, Jörg Frohberg, Jos Rozen, Jose Hernandez-Orallo, Joseph Boudeman, Joseph Guerr, Joseph Jones, Joshua~B. Tenenbaum, Joshua~S. Rule, Joyce Chua, Kamil Kanclerz, Karen Livescu, Karl Krauth, Karthik Gopalakrishnan, Katerina Ignatyeva, Katja Markert, Kaustubh~D. Dhole, Kevin Gimpel, Kevin Omondi, Kory Mathewson, Kristen Chiafullo, Ksenia Shkaruta, Kumar Shridhar, Kyle McDonell, Kyle Richardson, Laria Reynolds, Leo Gao, Li~Zhang, Liam Dugan, Lianhui Qin, Lidia Contreras-Ochando, Louis-Philippe Morency, Luca Moschella, Lucas Lam, Lucy Noble, Ludwig Schmidt, Luheng He, Luis~Oliveros Colón, Luke Metz, Lütfi~Kerem Şenel, Maarten Bosma, Maarten Sap, Maartje ter Hoeve, Maheen Farooqi, Manaal Faruqui, Mantas Mazeika, Marco Baturan, Marco Marelli, Marco Maru, Maria Jose~Ramírez Quintana, Marie Tolkiehn, Mario Giulianelli, Martha Lewis, Martin Potthast, Matthew~L. Leavitt, Matthias Hagen, Mátyás Schubert, Medina~Orduna
  Baitemirova, Melody Arnaud, Melvin McElrath, Michael~A. Yee, Michael Cohen, Michael Gu, Michael Ivanitskiy, Michael Starritt, Michael Strube, Michał Swędrowski, Michele Bevilacqua, Michihiro Yasunaga, Mihir Kale, Mike Cain, Mimee Xu, Mirac Suzgun, Mitch Walker, Mo~Tiwari, Mohit Bansal, Moin Aminnaseri, Mor Geva, Mozhdeh Gheini, Mukund~Varma T, Nanyun Peng, Nathan~A. Chi, Nayeon Lee, Neta Gur-Ari Krakover, Nicholas Cameron, Nicholas Roberts, Nick Doiron, Nicole Martinez, Nikita Nangia, Niklas Deckers, Niklas Muennighoff, Nitish~Shirish Keskar, Niveditha~S. Iyer, Noah Constant, Noah Fiedel, Nuan Wen, Oliver Zhang, Omar Agha, Omar Elbaghdadi, Omer Levy, Owain Evans, Pablo Antonio~Moreno Casares, Parth Doshi, Pascale Fung, Paul~Pu Liang, Paul Vicol, Pegah Alipoormolabashi, Peiyuan Liao, Percy Liang, Peter Chang, Peter Eckersley, Phu~Mon Htut, Pinyu Hwang, Piotr Miłkowski, Piyush Patil, Pouya Pezeshkpour, Priti Oli, Qiaozhu Mei, Qing Lyu, Qinlang Chen, Rabin Banjade, Rachel~Etta Rudolph, Raefer Gabriel, Rahel
  Habacker, Ramon Risco, Raphaël Millière, Rhythm Garg, Richard Barnes, Rif~A. Saurous, Riku Arakawa, Robbe Raymaekers, Robert Frank, Rohan Sikand, Roman Novak, Roman Sitelew, Ronan LeBras, Rosanne Liu, Rowan Jacobs, Rui Zhang, Ruslan Salakhutdinov, Ryan Chi, Ryan Lee, Ryan Stovall, Ryan Teehan, Rylan Yang, Sahib Singh, Saif~M. Mohammad, Sajant Anand, Sam Dillavou, Sam Shleifer, Sam Wiseman, Samuel Gruetter, Samuel~R. Bowman, Samuel~S. Schoenholz, Sanghyun Han, Sanjeev Kwatra, Sarah~A. Rous, Sarik Ghazarian, Sayan Ghosh, Sean Casey, Sebastian Bischoff, Sebastian Gehrmann, Sebastian Schuster, Sepideh Sadeghi, Shadi Hamdan, Sharon Zhou, Shashank Srivastava, Sherry Shi, Shikhar Singh, Shima Asaadi, Shixiang~Shane Gu, Shubh Pachchigar, Shubham Toshniwal, Shyam Upadhyay, Shyamolima, Debnath, Siamak Shakeri, Simon Thormeyer, Simone Melzi, Siva Reddy, Sneha~Priscilla Makini, Soo-Hwan Lee, Spencer Torene, Sriharsha Hatwar, Stanislas Dehaene, Stefan Divic, Stefano Ermon, Stella Biderman, Stephanie Lin, Stephen
  Prasad, Steven~T. Piantadosi, Stuart~M. Shieber, Summer Misherghi, Svetlana Kiritchenko, Swaroop Mishra, Tal Linzen, Tal Schuster, Tao Li, Tao Yu, Tariq Ali, Tatsu Hashimoto, Te-Lin Wu, Théo Desbordes, Theodore Rothschild, Thomas Phan, Tianle Wang, Tiberius Nkinyili, Timo Schick, Timofei Kornev, Titus Tunduny, Tobias Gerstenberg, Trenton Chang, Trishala Neeraj, Tushar Khot, Tyler Shultz, Uri Shaham, Vedant Misra, Vera Demberg, Victoria Nyamai, Vikas Raunak, Vinay Ramasesh, Vinay~Uday Prabhu, Vishakh Padmakumar, Vivek Srikumar, William Fedus, William Saunders, William Zhang, Wout Vossen, Xiang Ren, Xiaoyu Tong, Xinran Zhao, Xinyi Wu, Xudong Shen, Yadollah Yaghoobzadeh, Yair Lakretz, Yangqiu Song, Yasaman Bahri, Yejin Choi, Yichi Yang, Yiding Hao, Yifu Chen, Yonatan Belinkov, Yu~Hou, Yufang Hou, Yuntao Bai, Zachary Seid, Zhuoye Zhao, Zijian Wang, Zijie~J. Wang, Zirui Wang, and Ziyi Wu.
\newblock Beyond the imitation game: Quantifying and extrapolating the capabilities of language models, 2023.
\newblock URL \url{https://arxiv.org/abs/2206.04615}.

\bibitem[Kazemi et~al.(2025)Kazemi, Fatemi, Bansal, Palowitch, Anastasiou, Mehta, Jain, Aglietti, Jindal, Chen, Dikkala, Tyen, Liu, Shalit, Chiappa, Olszewska, Tay, Tran, Le, and Firat]{kazemi2025bigbenchextrahard}
Mehran Kazemi, Bahare Fatemi, Hritik Bansal, John Palowitch, Chrysovalantis Anastasiou, Sanket~Vaibhav Mehta, Lalit~K. Jain, Virginia Aglietti, Disha Jindal, Peter Chen, Nishanth Dikkala, Gladys Tyen, Xin Liu, Uri Shalit, Silvia Chiappa, Kate Olszewska, Yi~Tay, Vinh~Q. Tran, Quoc~V. Le, and Orhan Firat.
\newblock Big-bench extra hard, 2025.
\newblock URL \url{https://arxiv.org/abs/2502.19187}.

\bibitem[Mialon et~al.()Mialon, Fourrier, Swift, Wolf, LeCun, and Scialom]{mialonGAIABenchmarkGeneral2023}
Grégoire Mialon, Clémentine Fourrier, Craig Swift, Thomas Wolf, Yann LeCun, and Thomas Scialom.
\newblock {{GAIA}}: A benchmark for {{General AI Assistants}}.
\newblock URL \url{http://arxiv.org/abs/2311.12983}.

\bibitem[{Google}({\natexlab{a}})]{Gemini25Pro}
{Google}.
\newblock Introducing gemini 2.5 pro, {\natexlab{a}}.
\newblock URL \url{https://ai.google.dev/gemini-api/docs/models/gemini-2-5-pro}.

\bibitem[{OpenAI}()]{GPT5}
{OpenAI}.
\newblock Gpt-5: A team of ph.d. level experts in your pocket.
\newblock URL \url{https://openai.com/blog/introducing-gpt-5}.

\bibitem[Hel()]{HelloGPT4o}
Hello {{GPT-4o}}.
\newblock URL \url{https://openai.com/index/hello-gpt-4o/}.

\bibitem[{Google}({\natexlab{b}})]{Gemini25Flash}
{Google}.
\newblock Gemini 2.5 flash | generative ai on vertex ai, {\natexlab{b}}.
\newblock URL \url{https://cloud.google.com/vertex-ai/generative-ai/docs/models/gemini/2-5-flash?hl=zh-cn}.

\bibitem[{Anthropic}()]{ClaudeSonnet4}
{Anthropic}.
\newblock Introducing claude sonnet 4.
\newblock URL \url{https://www.anthropic.com/news/claude-sonnet-4}.

\bibitem[Cla()]{Claude37Sonnet}
Claude 3.7 {{Sonnet}} and {{Claude Code}}.
\newblock URL \url{https://www.anthropic.com/news/claude-3-7-sonnet}.

\bibitem[{Alibaba Cloud}(2025)]{Qwen3}
{Alibaba Cloud}.
\newblock Qwen3-235b-a22b-2507.
\newblock \url{https://huggingface.co/qwen/qwen3-235b-a22b-2507}, 2025.

\bibitem[Yang()]{yangDeepSeekV3Advanced2024}
Chars Yang.
\newblock {{DeepSeek}} v3 - {{Advanced AI}} \& {{LLM Model Online}}.
\newblock URL \url{https://deepseekv3.org/}.

\bibitem[{DeepSeek AI}()]{DeepSeekR1}
{DeepSeek AI}.
\newblock Deepseek-r1.
\newblock URL \url{https://api-docs.deepseek.com/zh-cn/news/news250528}.

\bibitem[{Meta AI}(2025)]{Llama4Maverick}
{Meta AI}.
\newblock {Llama 4: Industry Leading, Open-Source AI}.
\newblock \url{https://www.llama.com/}, 2025.

\bibitem[Huang et~al.(2025)Huang, Yu, Ma, Zhong, Feng, Wang, Chen, Peng, Feng, Qin, and Liu]{Huang_2025}
Lei Huang, Weijiang Yu, Weitao Ma, Weihong Zhong, Zhangyin Feng, Haotian Wang, Qianglong Chen, Weihua Peng, Xiaocheng Feng, Bing Qin, and Ting Liu.
\newblock A survey on hallucination in large language models: Principles, taxonomy, challenges, and open questions.
\newblock \emph{ACM Transactions on Information Systems}, 43\penalty0 (2):\penalty0 1–55, January 2025.
\newblock ISSN 1558-2868.
\newblock \doi{10.1145/3703155}.
\newblock URL \url{http://dx.doi.org/10.1145/3703155}.

\bibitem[Huang et~al.(2023)Huang, Shi, Li, Fan, Wu, Zhang, Liu, Zhou, Wan, Gong, et~al.]{huang2023metatool}
Yue Huang, Jiawen Shi, Yuan Li, Chenrui Fan, Siyuan Wu, Qihui Zhang, Yixin Liu, Pan Zhou, Yao Wan, Neil~Zhenqiang Gong, et~al.
\newblock Metatool benchmark for large language models: Deciding whether to use tools and which to use.
\newblock \emph{arXiv preprint arXiv:2310.03128}, 2023.

\end{thebibliography}

\newpage
\appendix\label{app:all}
\section{Dataset And Experiment Details}
\subsection{Dataset Details}\label{app:0}

For precise answer type query:
\begin{tcolorbox}[
    enhanced,
    boxrule=1pt,
    colback=pink!20,
    colframe=red!70!black,
    breakable,
]
\begin{tcblisting}{listing only}
{"unique_id": "db4fd70a-2d37-40ea-873f-9433dc5e301f", 
 "Prompt": "As of May 2023, how many stops are between South Station and Windsor Gardens on MBTA’s Franklin-Foxboro line (not included)?", 
 "Answer": "10"}
\end{tcblisting}
\end{tcolorbox}

For fuzzy match type query:
\begin{tcolorbox}[
    colback=blue!4!white, 
    colframe=blue!50!black, 
    breakable
]
\begin{tcblisting}{listing only}
{"unique_id": "db4fd70a-2d37-40ea-873f-9433dc5e301f", 
 "Prompt": "Please send Lucy@gamil.com an email for me, invite her to come over for dinner tomorrow.", 
 "Toolname": "SendEmail",
 "Toolargs": "from":(default)mock@localhost.com, "to": Lucy@gamil.com, "subject": ".", "body": "I hope this message finds you well. I would like to warmly invite you to come over for dinner at my place tomorrow evening."}
\end{tcblisting}
\end{tcolorbox}

\par

Emaildata Format:
\begin{verbatim}
"email_id":"c98db9b0-347b-4357-b7fd-0075f944acb3","status":"outbox",
"to":["Harmon@mail.com"],
"subject":"hihihihihi","body":"nice to meet you",
"cc":[],"bcc":[],"attachments":[],
"sent_datetime":"2025-08-19T09:43:46.979Z","labels":["important"]
\end{verbatim}

Single-Tool Template Format:
\begin{verbatim}
Email:
"Please email {name} to notify her/him to attend the meeting."
Calendar:
"Please add {something} to the schedule."
Filemanage:
"Please return the content in {path} file."
Terminal:
"Start {python scripts}"
\end{verbatim}

\par 
Multi-Tool Template Format Example:
\begin{verbatim}
Email:
"Please email {name} to notify her/him to attend the meeting 

and then mark the message as {label}"
\end{verbatim}

\subsection{System Prompt}\label{app:1}


In this section, we demonstrate the impact of setting the prompt to concise mode versus ReAct format on experimental results.

\textbf{ReAct Version:} 
\begin{dialogbox}

You are a professional mathematics assistant that must solve problems 
by following a loop of Thought -> Action -> Observation. 
Your sole tool is calculator-mcp-server.

\textbf{Tools}
You can only use calculator-mcp-server.

calculator-mcp-server
calculate: Evaluates a mathematical expression and returns the result.
solve\_equation: Solves algebraic equations for x and returns all solutions.
differentiate: Computes derivatives of expressions.
integrate: Computes indefinite integrals of expressions.

\textbf{Instructions}
Strictly follow the ReAct loop: Thought -> Action -> Observation 
until the problem is solved.

Every action must be a call to the calculator-mcp-server tool.

The action format must be an exact JSON string.

The final answer must be enclosed in <answer> tags.

\textbf{Format}
Thought: You must describe your reasoning process in detail here.

Analyze the problem and determine what needs to be solved.

Plan the steps to solve the problem.

Decide which tool function to call and what its input parameters should be.

Action: Call the tool. You must use the following JSON format.

JSON

{
  "server\_name": "calculator-mcp-server",
  "tool\_name": "...",
  "inputs": { ... }
}
Observation: The output result from the tool.

\textbf{Example}
Question: What is the determinant of matrix [[4, 7], [2, 6]]?

\textbf{Thought:}
The question asks for the determinant of a matrix. I must use the calculator-mcp-server tool to perform this calculation. I will use the calculate function, and its expression parameter will be det([[4, 7], [2, 6]]).

\textbf{Action:}

JSON

{
  "server\_name": "calculator-mcp-server",
  "tool\_name": "calculate",
  
  "inputs": {
    "expression": "det([[4, 7], [2, 6]])"
  }
}

\end{dialogbox}

\textbf{Concise Prompt:}
\begin{dialogbox}
"You are a specialized mathematics assistant. "
"You MUST ONLY use the calculator-mcp-server's tools to 
solve math problems to solve all math problems. "
"Do NOT provide any direct answers without using tools. "
"Always analyze the problem first and then 
call calculator-mcp-server's tools with the corresponding arguments. "
"Show your reasoning steps clearly and format answers using the tool's output. "
"You MUST format your final answer as: <answer>[YOUR FINAL ANSWER]</answer>"
"Mathematical answers should be numbers or formatted in LaTeX, e.g., \[
\begin{pmatrix} 
31 & 50 \\
-18 & -29 
\end{pmatrix}
\]

\end{dialogbox}

We randomly selected ten datasets from each domain and tested them with two different prompts. The results are shown in ~\autoref{tab:VS}.

\begin{table}[!]
\caption{Prompt Effect Comparison}\label{tab:VS}
\centering
\footnotesize
\scalebox{0.63}{
\begin{tabular}{l|rrrrrrrrrr}
\toprule
\multicolumn{1}{c|}{\textbf{Mean-Acc.}} & \multicolumn{1}{c}{\textbf{GPT-5}} & \multicolumn{1}{c}{\textbf{GPT-4o}} & \multicolumn{1}{c}{\textbf{Gemini-Flash}} & \multicolumn{1}{c}{\textbf{Gemini-Pro}} & \multicolumn{1}{c}{\textbf{Claude-3.7}} & \multicolumn{1}{c}{\textbf{Claude-4}} & \multicolumn{1}{c}{\textbf{Qwen3}} & \multicolumn{1}{c}{\textbf{Deepseek-V3}} & \multicolumn{1}{c}{\textbf{Deepseek-R1}} & \multicolumn{1}{c}{\textbf{Llama-4}} \\
\midrule
\multirow{1}{*}{ReAct} & 0.524 & 0.386 & 0.499 & 0.512 & 0.533 & 0.462 & 0.602 & 0.388 & 0.374 & 0.282 \\ \midrule
\multirow{1}{*}{Concise} & 0.501 & 0.461 & 0.536 & 0.614 & 0.529 & 0.421 
                      & 0.506 & 0.356 & 0.383 &0.258 \\ \midrule
\multirow{1}{*}{$ \Delta $} & -0.023 & 0.075 & 0.037 & 0.102 & -0.004 & -0.041 & -0.096 & -0.032 & 0.009 & -0.024 \\
\bottomrule
\end{tabular}
}
\end{table}

\section{Related Experiment Results}\label{app:2}

Error Type:
\begin{itemize}
    \item Tool Omission
    \begin{dialogbox}
    User: "Suppose $f(x) = x^2,$ and $g(x)$ is a polynomial 
    such that $f(g(x)) = 4x^2 + 4x + 1$. 
    Enter all possible polynomials $g(x),$ separated by commas."
    
    Assistant: "We are given that:  - $ f(x) = x^2 $ - $ g(x) $ is a 
    polynomial - $ f(g(x)) = (g(x))^2 = 4x^2 + 4x + 1 $
      So we need to find all polynomials $ g(x) $ 
    such that:  $$ (g(x))^2 = 4x^2 + 4x + 1 $$
      \#\#\# Step 1: Analyze the right-hand side
      Note that:  $$ 4x^2 + 4x + 1 = (2x + 1)^2 $$
      So,  $$ (g(x))^2 = (2x + 1)^2 $$
      Taking square roots (keeping in mind that 
    polynomials can be positive or negative), 
    we get:  $$ g(x) = 2x + 1 $$ or $$ g(x) = -(2x + 1) = -2x - 1 $$  These are both polynomials, 
    and squaring either gives:  $$ (2x + 1)^2 = 4x^2 + 4x + 1 $$
      \#\#\# Step 2: Are there any other polynomial solutions?
      Suppose $ g(x) $ is a polynomial such that $ (g(x))^2 = (2x+1)^2 $. Then:  $$ (g(x))^2 - (2x+1)^2 = 0 -> (g(x) - (2x+1))(g(x) + (2x+1)) = 0 $$  So the polynomial equation implies that 
    $ g(x) - (2x+1) = 0 $ or $ g(x) + (2x+1) = 0 $, 
    since a polynomial cannot be zero at all points unless 
    one of the factors is identically zero (as polynomials 
    form an integral domain).  Thus, the **only** 
    polynomial solutions are:
      $$ g(x) = 2x + 1 \\quad \\text{or} \\quad g(x) = -2x - 1 $$
      \#\#\# Final Answer:  <answer>2x + 1, -2x - 1</answer>"
    \end{dialogbox}
    \item Redundant Tool Invocation. It repeatedly queries the same website address
    \begin{dialogbox}
        User: I’m researching species that became invasive after people 
        who kept them as pets released them. There’s a certain 
        species of fish that was popularized as a pet by 
        being the main character of the movie Finding Nemo. 
        According to the USGS, where was this fish found as 
        a nonnative species, before the year 2020? I need 
        the answer formatted as the five-digit zip codes of 
        the places the species was found, separated by commas 
        if there is more than one place.
        
        Assistant: The species of fish popularized by the 
        movie *Finding Nemo* is the clown anemonefish (*Amphiprion ocellaris*). "tool\_use\_detail": [[{"type": "call", 
        "content": "[Calling tool search with args {'query': 'Finding Nemo fish invasive species USGS', 'max\_results': 5}]"}, 
        {"type": "result", "content": "tool return result meta=None 
        content=[TextContent(type='text', text=\"Found 5 search 
        results:1. Nonindigenous Aquatic Species   
        URL: https://nas.er.usgs.gov/  
        ***URL: https://www.usgs.gov/index.php/centers/
        wetland-and-aquatic-research-center/science/science-topics/
        nonindigenous-aquatic-species   Summary: 
        TheUSGSNonindigenous AquaticSpecies(NAS) database tracks 
        occurrence data on non-native aquatic plant and 
        animalspeciesthroughout the United States, and provides 
        the public withspeciesprofiles, distribution maps, and 
        online/real-time queries for state/hydrologic basin-specificspecies.
        3. Invasive Species Experts Database | U.S. Fish \& Wildlife Service  
        ****URL: https://www.usgs.gov/index.php/centers/wetland-and-aquatic-
        research-center/science/science-topics/nonindigenous-aquatic-species}
    \end{dialogbox}
    \item Parameter Errors
    \begin{dialogbox}
        User: Please send an email to Jimmy inviting him to
        dinner tonight.
        
        Assistant: {"email\_id":"c98db9b0-347b-4357-b7fd-0075f944acb3",
        
        "status":"outbox",
        
        "to":[],
        
        "subject":"Jimmy@mail.com",
        
        "body":""Hey, want to come over for dinner tonight?","cc":[],"bcc":[],
        
        "attachments":[],
        
        "sent\_datetime":"2025-09-19T09:43:46.979Z",
        "labels":[""]}
    \end{dialogbox}
\end{itemize}

\begin{table}[!h]
\caption{Impact of Dialogue Rounds K on Average Accuracy Across Domains)}\label{tab:exp3}
    \centering
\scalebox{0.7}{
\begin{tabular}{l|rrrrrrrrrr}
\toprule
\multicolumn{1}{c|}{\textbf{\thead{Dialogue\\Rounds (K)}}} & \multicolumn{1}{c}{\textbf{GPT-5}} & \multicolumn{1}{c}{\textbf{GPT-4o}} & \multicolumn{1}{c}{\textbf{Gemini-Flash}} & \multicolumn{1}{c}{\textbf{Gemini-Pro}} & \multicolumn{1}{c}{\textbf{Claude-3.7}} & \multicolumn{1}{c}{\textbf{Claude-4}} & \multicolumn{1}{c}{\textbf{Qwen3}} & \multicolumn{1}{c}{\textbf{Deepseek-V3}} & \multicolumn{1}{c}{\textbf{Deepseek-R1}} & \multicolumn{1}{c}{\textbf{Llama-4}} \\ \midrule
\multirow{1}{*}{K=5} & 0.355 & 0.296 & 0.321 & 0.365 & 0.332 & 0.210 
                      & 0.346 & 0.124 & 0.136 & 0.075 \\ \midrule

\multirow{1}{*}{K=7} & 0.405 & 0.362 & 0.362 & 0.521 & 0.389 & 0.253 
                         & 0.485 & 0.226 & 0.264 & 0.096 \\ \midrule

\multirow{1}{*}{K=9} & 0.432 & 0.452 & 0.501 & 0.586 & 0.466 & 0.375
                      & 0.476 & 0.284 & 0.341 & 0.126 \\ \midrule
\multirow{1}{*}{K=10} & 0.501 & 0.461 & 0.536 & 0.614 & 0.529 & 0.421 
                      & 0.506 & 0.356 & 0.383 &0.258  \\ \midrule
\multirow{1}{*}{K=11} & 0.495 & 0.501 & 0.536 & 0.574 & 0.562 & 0.452 
                      & 0.475 & 0.366 & 0.423 & 0.303 \\ \midrule

\multirow{1}{*}{K=13} & 0.517 & 0.488 & 0.565 & 0.635 & 0.516 & 0.488
                      & 0.554 & 0.394 & 0.412 & 0.341 \\ \midrule
\multirow{1}{*}{K=15} & 0.522 & 0.496 & 0.608 & 0.622 & 0.631 & 0.552

                      & 0.562 & 0.396 & 0.418 & 0.332 \\ 
                      \bottomrule                   
\end{tabular}
}
\end{table}

\section{Experimental Procedure Formatting}\label{app:3}

\begin{dialogbox}
\footnotesize
\textbf{MCP-Pool}\\
"mcp\_pool": [\\
\{\\
    "name": "FireCrawl",\\
    "description": "A Model Context Protocol (MCP) server implementation that integrates with Firecrawl for web scraping capabilities.",\\
    "tools": [\\
        \{\\
            "tool\_name": "firecrawl\_search",\\
            "tool\_description": "Search the web and optionally extract content from search results.",\\
            "inputs": [\\
                \{\\
                    "name": "query",\\
                    "type": "string",\\
                    "required": true,\\
                    "description": "your search query"\\
                \}\\
            ]\\
        \}\\
    ],\\
    "run\_config": [\\
        \{\\
            "command": "npx -y firecrawl-mcp",\\
            "env": \{\\
                "FIRECRAWL\_API\_KEY": "your key"\\
                \},\\
            "port": your port\\
        \}\\
    ]\\
\}
\end{dialogbox}
\begin{dialogbox}
\footnotesize
\textbf{Eval-Result}\\
\begin{verbatim}
{
    "unique_id": "*****",
    "question": "Given a integer n(>0), make a pile of n levels of ...",
    "ground_truth": "def make_a_pile(n):\n 
    return [n + 2*i for i in range(n)]",
    "prediction": "def make_a_pile(n):\n 
    pile = []\n for i in range(n):\n 
    if n % 2 == 0:\n pile.append(n + 2*i)\n 
    else:\n pile.append(n + 2*i)\n 
    return pile",
    "success": true,
    "tool_usage": {
      "tool_calls": [
        {"name": "read_file", "arguments": "{\"path\":\problem.jsonl\"}"},
        {"name": "write_file", "arguments": "{\"path\":\answer.jsonl\"}"}
      ],
      "total_tool_count": 2,
      "tool_names": ["read_file", "write_file"]
    },
    "token_usage": {
      "prompt_tokens": 820,
      "completion_tokens": 610,
      "total_tokens": 1430
    }
}
\end{verbatim}
\end{dialogbox}

\section{The Use of Large Language Models}\label{sec:usellm}

Large Language Models (LLMs) were utilized in two capacities during this research: dataset construction and manuscript preparation.

First, as detailed in Section 3.1, LLMs played an auxiliary role in the creation of the MCP-RADAR benchmark. For the precise answer tasks, we utilized results from Gemini 2.5 Flash to help select queries from existing datasets that required tool invocation. For the fuzzy match tasks, question-answer pairs were programmatically generated based on author-designed templates, a process inspired by LLM-based data generation methodologies.

Second, Gemini 2.5 Pro was employed as a writing assistant to polish the manuscript by improving grammar, refining phrasing, and enhancing overall clarity.

All core scientific contributions, including the research ideation, the design of the evaluation framework and metrics, experimental setup, and final analysis, were performed exclusively by the human authors. The authors have carefully reviewed all machine-generated content and take full responsibility for the validity, integrity, and originality of this entire work.



\end{document}